\documentclass[journal]{IEEEtran}
\usepackage{cite}
\usepackage{amsmath,amsfonts}
\usepackage{graphicx}
\usepackage{array}
\usepackage[caption=false,font=footnotesize]{subfig}
\usepackage{url}
\usepackage{color,soul}
\usepackage{multirow}
\usepackage[mdyy,hhmmss,12hr]{datetime}
\usepackage{booktabs}
\usepackage{hyperref}
\usepackage{algorithm}
\usepackage[noend]{algpseudocode}
\usepackage{amssymb,amsmath,amsthm}
\newtheorem{lemma}{Lemma}

\definecolor{shapecolor}{rgb}{0.0,0.5,0.0}

\graphicspath{{./fig/}{./fig/scatter/}}

\begin{document}
\title{Text-controlled Motion Mamba: Text-Instructed Temporal Grounding of Human Motion}
\author{Xinghan~Wang,
Zixi~Kang
and~Yadong~Mu
\thanks{Xinghan Wang, Zixi Kang and Yadong Mu are with the Wangxuan Institute of Computer Technology, Peking University, Beijing 100871, China (e-mail: xinghan\_wang@pku.edu.cn, forever.kzx0713@stu.pku.edu.cn, myd@pku.edu.cn). Yadong Mu is the corresponding author. The research is supported by a grant from Bytedance (No.CT20250811106734).}
}

\maketitle

\begin{abstract}
Human motion understanding is a fundamental task with diverse practical applications, facilitated by the availability of large-scale motion capture datasets. Recent studies focus on text-motion tasks, such as text-based motion generation, editing and question answering. In this study, we introduce the novel task of text-based human motion grounding (THMG), aimed at precisely localizing temporal segments corresponding to given textual descriptions within untrimmed motion sequences. Capturing global temporal information is crucial for the THMG task. However, Transformer-based models that rely on global temporal self-attention face challenges when handling long untrimmed sequences due to the quadratic computational cost. We address these challenges by proposing Text-controlled Motion Mamba (TM-Mamba), a unified model that integrates temporal global context, language query control, and spatial graph topology with only linear memory cost. The core of the model is a text-controlled selection mechanism which dynamically incorporates global temporal information based on text query. The model is further enhanced to be topology-aware through the integration of relational embeddings. For evaluation, we introduce BABEL-Grounding, the first text-motion dataset that provides detailed textual descriptions of human actions along with their corresponding temporal segments. Extensive evaluations demonstrate the effectiveness of TM-Mamba on BABEL-Grounding.
\end{abstract}


\section{Introduction}
Human motion understanding is a crucial task with a wide range of applications. Recent years have witnessed the flourishing of large-scale motion capture databases~\cite{mahmood2019amass, shahroudy2016ntu, mandery2015kit, ionescu2013human3}, which greatly facilitate the end-to-end training of various motion-related tasks, such as motion prediction~\cite{li2021multiscale, li2020multitask, wang2021pvred, wang2023dynamic}, action recognition~\cite{guo2023b2c, wang2023neural, wang2024localized, myung2024degcn, li2022smam, zhu2022multilevel, shi2020skeleton, yang2021feedback, hao2021hypergraph, bian2021structural, cheng2021extremely, wang2022contrast}, motion segmentation~\cite{xia2017human, yang2023lac}, etc. Based on these databases, recent studies~\cite{plappert2016kit, guo2022generating, liang2023intergen, han2023amd, tang2023flag3d, li2023sequential} have augmented motion datasets with textual annotations. These annotated motion-text pairs enable a range of text-motion tasks that require a joint understanding of human motion and language, such as text-based motion generation~\cite{tevet2022motionclip, petrovich2022temos, zhang2023t2m, zhang2024motiondiffuse}, motion captioning~\cite{guo2022tm2t, jiang2023motiongpt}, text-based motion editing~\cite{goel2023iterative, kim2023flame, zhang2023finemogen}, and motion question answering~\cite{endo2023motion}. However, in real-world scenarios, semantic actions often occur sparsely within lengthy motion sequences. Thus, a textual description of particular human actions often corresponds to specific temporal segments of the sequence rather than the entire sequence. 
Precisely localizing the temporal segments with text query presents a substantial challenge, which has broad applications across domains like surveillance, sports analytics, healthcare and autonomous driving.

In this study, we introduce the task of text-based human motion grounding (THMG) for the first time, which aims to identify the start and end timestamps of all segments corresponding to a given textual description from an untrimmed motion sequence.  

\begin{figure}[t] 
  \centering
  \includegraphics[width=\linewidth]{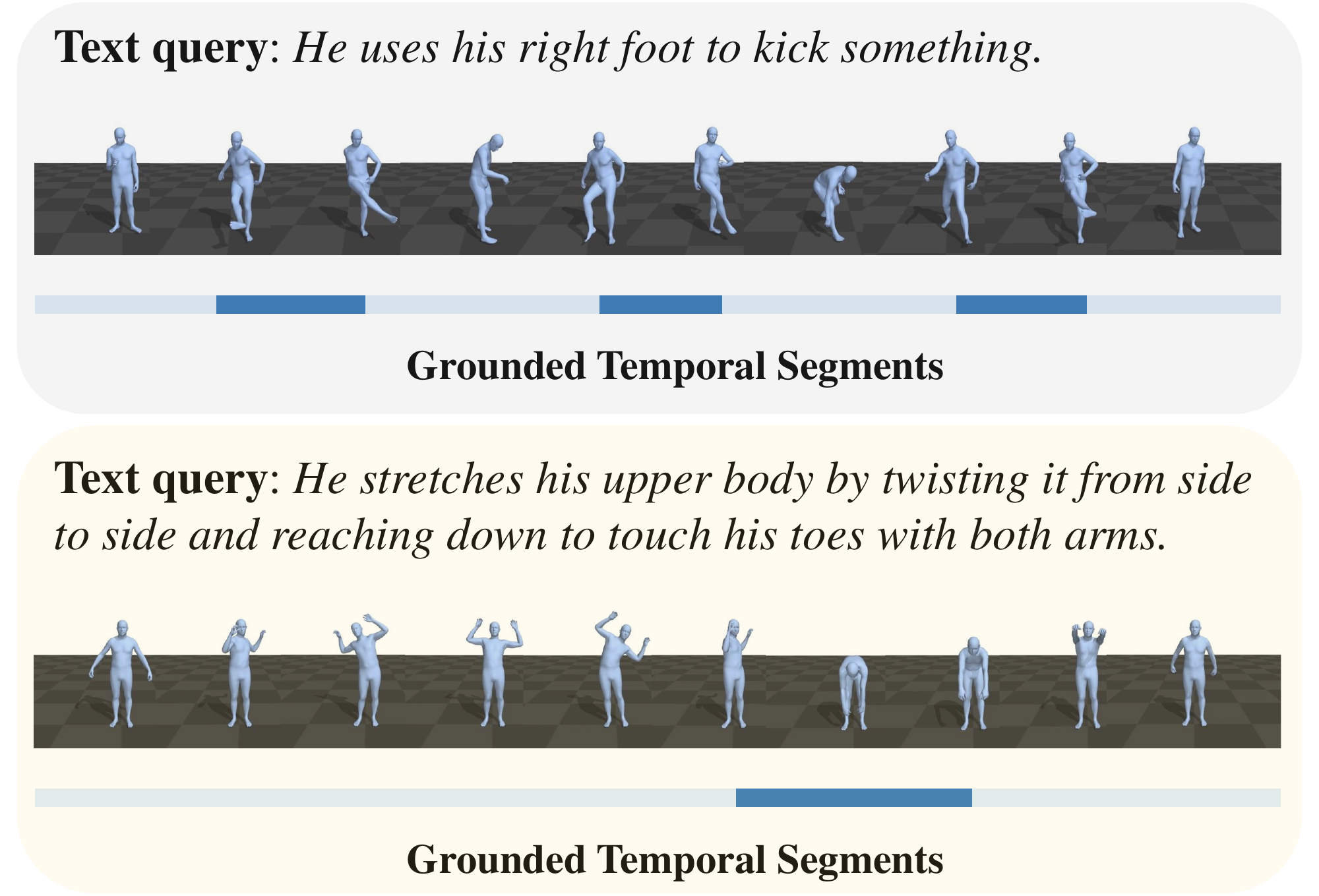}
  \caption{Illustration of the Text-based Human Motion Grounding (TMHG) task and samples of the proposed BABEL-Grounding dataset. Best viewed in color.} 
  \label{fig:motivation}
\end{figure}

A closely related task is video-based temporal grounding~\cite{lei2021detecting, yang2022tubedetr, lin2023univtg, mun2020local, zhang2020does} or video-based action localization~\cite{paul2018w, liu2019completeness, shou2017cdc, liu2024stepwise, zhang2022temporal, zhang2022actionformer, xu2021videoclip}, which seek to identify the temporal segments within an RGB video that correspond to a given textual description or action class. 
Yet, in application like gaming, AR, robotics, the RGB modality is unavailable, and data is inherently represented as 3D motion. Also, RGB and 3D motion capture different aspects of human activity: RGB encodes visual appearance, while 3D motion represents kinematics (joints, velocities, rotations), which offers unique benefits like robustness to background, privacy preservation, viewpoint invariance, and lightweight due to reduced dimensionality. 
Unlike existing motion temporal action localization tasks~\cite{yu2023frame}, the queries in THMG consist of arbitrary natural languages rather than a predefined set of action labels. As shown in Figure~\ref{fig:motivation}, the THMG task is highly challenging as it requires simultaneous consideration of several critical factors: (1) Achieving precise grounding of the time interval corresponding to the query within a long sequence demands the model's ability to grasp global temporal context effectively. (2) The model should jointly tackle the information from motion and language, ensuring a thorough fusion and interaction between the two modalities. (3) As human pose representation inherently possesses a graph structure, the model needs to capture the latent spatial topological information. 

Effectively capturing global temporal information is crucial for THMG task. Existing frameworks for text-motion analysis mostly adopt recurrent neural networks~\cite{guo2020action2motion, wang2022humanise} or temporal convolutions~\cite{zhang2023t2m, zhou2023avatargpt, jiang2023motiongpt, guo2022tm2t,guo2022generating, zhong2023attt2m} as the main workhorse. However, their ability to capture long-term dependency is quite limited. Recently, there has been a surge of interest in Transformer-based models~\cite{petrovich2022temos, tevet2022motionclip, zhai2023language} for modeling temporal dependencies. In these approaches, a global temporal self-attention mechanism is employed across all frames in the sequence. Nonetheless, these methods encounter challenges when dealing with untrimmed sequences that are very long, as computing a global temporal self-attention is exceedingly computationally expensive in such scenarios.


In this work, we aim to attack all the aforementioned challenges through a unified model that seamlessly incorporates temporal global context, language query control, and spatial graph topology. Our primary source of inspiration stems from the recently proposed state space model called Mamba~\cite{gu2023mamba}, an efficient model for handling long-term dependencies within lengthy sequences, while maintaining linear computational cost. Mamba has demonstrated its power across diverse domains, including language modeling and visual understanding. However, its potential application in human motion tasks remains largely unexplored.

The core of Mamba is an input-dependent selection mechanism, enabling the model to selectively propagate or forget information over time depending on the current input. This innovative design greatly enhances traditional State Space Model (SSM) methods, as Mamba is capable of grasping the global context of long sequences while filtering out irrelevant information. However, in the THMG task, the model must select information based textual queries as well. To address this challenge, a text-controlled selection mechanism is introduced, wherein the key idea is to condition the state transition matrix on both motion and text queries. Unlike existing multimodal Mamba methods~\cite{qiao2024vl, zhao2024cobra} that merely concatenate the textual and visual features and feed it into Mamba blocks, our approach is the first work in enabling texts to dictate the selective propagation of input information. This ensures that the model dynamically adjusts its focus based on the interplay between motion and text inputs. Furthermore, as human motion sequences inherently possess a graph topology, the original Mamba model is not suitable as it is designed to operate on univariate time series. To address this, we enhance Mamba by integrating relational information via graph neural networks into its state representation to facilitate topology awareness. The resulting framework, termed Text-Controlled Motion Mamba (TM-Mamba), can selectively extract relevant global context information based on textual queries in the motion sequence. 

For evaluation, existing datasets~\cite{punnakkal2021babel, lin2023motion} with frame-level temporal annotations are not directly applicable to the THMG task. While BABEL~\cite{punnakkal2021babel} provides temporal boundaries for all actions that occur in the sequence, it lacks complete detailed textual descriptions, offering only simple categorical phrases. Motion-X~\cite{lin2023motion} offers detailed pose descriptions for each frame, but its texts focus on the movement of human body parts at each timestep, making it incapable of establishing a mapping from semantic action descriptions to temporal segments. To address this gap, based on BABEL, a new dataset called BABEL-Grounding is introduced to serve as a benchmark for evaluating THMG task. BABEL-Grounding is the first text-motion dataset that provides detailed textual descriptions of human actions along with their corresponding temporal segments in untrimmed motion sequences. Like real-world scenarios, each query may correspond to multiple temporal segments. Extensive evaluations of TM-Mamba are conducted on the newly introduced dataset, demonstrating its effectiveness in THMG task. Our primary contributions can be summarized as follows:
\begin{itemize}
    \item We introduce a new text-motion task, text-based human motion grounding (THMG), along with a text-motion dataset called BABEL-Grounding tailored specifically for THMG, which is the first of its kind.
    \item We proposed TM-Mamba, a unified model with only linear memory cost specially crafted for THMG task, which is the first work that incorporates a text-controlled selection mechanism into the Mamba framework.
    \item Extensive evaluations on the BABEL-Grounding dataset demonstrates the efficacy of the proposed TM-Mamba.
\end{itemize}

\section{Related work}
\subsection{Datasets for Text-Motion Learning}
This section presents an overview of existing human motion datasets annotated with texts. These text-motion datasets are primarily developed for text-driven motion generation task, hence they typically include textual descriptions at the sequence level for each motion sequence. For example, the KIT Motion Language dataset~\cite{plappert2016kit} is the first that provides human motion alongside corresponding sequence-level textual descriptions. Following this direction, several subsequent works have endeavored to construct larger-scale datasets of similar kind, such as HumanML3D~\cite{guo2022generating}, InterHuman~\cite{liang2023intergen}, HumanLong3D~\cite{han2023amd}, FLAG3D~\cite{tang2023flag3d}, STDM~\cite{li2023sequential}. Some datasets go beyond mere textual annotations and incorporate additional contextual information. For instance, HUMANISE~\cite{wang2022humanise} integrates 3D scene information to facilitate motion generation within 3D environments, while HOI-Diff~\cite{peng2023hoi} incorporates object geometry information to support human-object interaction during generation. Furthermore, there are datasets tailored for diverse text-motion tasks beyond generation alone. Examples include PoseScript~\cite{delmas2022posescript}, which focuses on static pose generation and pose captioning, and PoseFix~\cite{delmas2023posefix}, which targets motion editing tasks.

However, all of the aforementioned datasets only contain sequence-level annotations, rendering them unsuitable for temporal tasks. To address this limitation, various efforts have been made to develop motion datasets with specific temporal information. For instance, BABEL~\cite{punnakkal2021babel} introduces a motion dataset with frame-wise annotations, providing temporal spans for each action label, thereby enabling tasks such as action localization. Constructed upon BABEL, BABEL-QA~\cite{endo2023motion} extends the dataset by incorporating question-answer pairs, aiming to facilitate motion-based question answering. Another example is HuMMan-MoGen~\cite{zhang2023finemogen}, which is built upon the HuMMan~\cite{cai2022humman} dataset, where each sequence is divided into predefined action phases, along with phase-level detailed annotations describing the movement of each body part. Recently, Motion-X~\cite{lin2023motion} offers part-level textual annotations for human pose at each frame. However, the annotations of Motion-X are based on individual poses and lack a mapping from textual descriptions of semantic actions to temporal boundaries.

\subsection{Text-Motion Multi-modal Learning}
Recently, there has been a growing interest in text-motion multi-modal learning. Current research mainly focuses on text-to-motion task (also known as text-driven motion generation), where human motion sequences are generated based on natural language~\cite{ahuja2019language2pose, ghosh2021synthesis,  yuan2023physdiff, ma2022pretrained, jin2023act,zhang2023finemogen,azadi2023make, wang2023fg,kong2023priority, karunratanakul2023guided, qian2023breaking, li2023sequential, zhong2023attt2m, tevet2022motionclip, petrovich2022temos, chen2023executing, zhou2023ude, zhang2023t2m, lin2023being, yang2023synthesizing, zhai2023language, han2023amd, han2023hutumotion, hoang2024motionmix, xie2023towards, zhang2023motiongpt, shafir2023human, wei2023nerm, xie2023omnicontrol, wang2022humanise, guo2022generating, athanasiou2022teach}. The key challenge is to learn a joint embedding space for motion and text. For instance, MotionCLIP~\cite{tevet2022motionclip} aligns human motion embeddings with CLIP space through cosine similarity loss to inherit the semantic structure of CLIP latent space. A traditional Transformer-based auto-encoder is employed to generate motion sequences. TEMOS~\cite{petrovich2022temos} addresses stochastic motion generation using a VAE structure, aligning the embedding space of motion and text via KL-divergence between their latent distributions. Other works focus on employing various modern generation methods for improved conditional generation. For instance, T2M-GPT~\cite{zhang2023t2m} utilizes Vector Quantised Variational AutoEncoder (VQ-VAE) to encode motion sequences into discrete tokens, enabling GPT-like autoregressive generation and training through next-token prediction, with text embedding serving as prior. MotionDiffuse~\cite{zhang2024motiondiffuse} incorporates Denoising Diffusion Probabilistic Models (DDPM) into the task of text-motion generation,  greatly enhancing the diversity and fidelity of generated sequences.

In recent years, some works~\cite{dabral2023mofusion, guo2022tm2t, jiang2023motiongpt, zhou2023avatargpt, kim2023flame} aim for more general text-motion models capable of seamlessly handling various text-motion tasks simultaneously. For example, TM2T~\cite{guo2022tm2t} establishes a bi-modal mutual mapping between texts and tokenized human motion using autoregressive neural machine translators (NMT), effectively handling both text-to-motion and motion-to-text tasks. MotionGPT~\cite{jiang2023motiongpt} utilizes VQ-VAE to create a motion tokenizer and vocabulary, and then perform pretraining on both motion and text in a unified manner, by treating human motion as a foreign language. 
Additionally, some works explore various text-motion tasks beyond generation, including motion question answering~\cite{endo2023motion}, text-based motion editing~\cite{goel2023iterative, kim2023flame, zhang2023finemogen}, text-motion retrieval~\cite{petrovich2023tmr, messina2023text}, etc.

\subsection{State Space Model}
State space models (SSMs) are a series of sequential models renowned for their computation and memory efficiency and the ability to model long-term dependencies. The pioneering work, S4~\cite{gu2021efficiently}, first proposed applying HiPPO~\cite{gu2020hippo} initialization to enable SSMs to maintain long-range memory. Subsequent studies~\cite{smith2022simplified, fu2022hungry, mehta2022long, gu2022parameterization, gupta2022diagonal, hasani2022liquid, gu2022train} have followed this direction, further improving the space structure and network architecture of S4.

Recently, Mamba~\cite{gu2023mamba, dao2024transformers} introduces an input-dependent selection mechanism into SSMs, demonstrating linear time efficiency in long-sequence modeling and achieving outstanding performance across various sequential tasks. The model has been adapted to diverse tasks, including image restoration~\cite{guo2024mambair}, image segmentation~\cite{wang2024semi, ruan2024vm, ma2024u, xing2024segmamba, liu2024swin}, point cloud~\cite{liang2024pointmamba}, video understanding~\cite{yang2024vivim, li2024videomamba, chen2024video}, pan-sharpening~\cite{he2024pan}, graph analysis~\cite{behrouz2024graph, wang2024graph}, multimodal learning~\cite{qiao2024vl, zhao2024cobra}. There're also some efforts~\cite{zhu2024vision, liu2024vmamba} aiming to establish a universal visual backbone based on Mamba by sequentializing images using patch-based methods akin to ViT. In a more recent study, Motion Mamba~\cite{zhang2024motion} presents a motion generation model that incorporates the Mamba block into a denoising U-Net architecture, effectively enhancing motion consistency across frames. In contrast to our method which deeply integrates text control into the selection mechanism, their model’s selective state space component remains unimodal, employed solely for spatial-temporal feature extraction of the motion sequence, while textual information is integrated during the diffusion process via a conditional denoiser.

\begin{figure}[t]
\begin{center}
\includegraphics[width=1.0\linewidth]{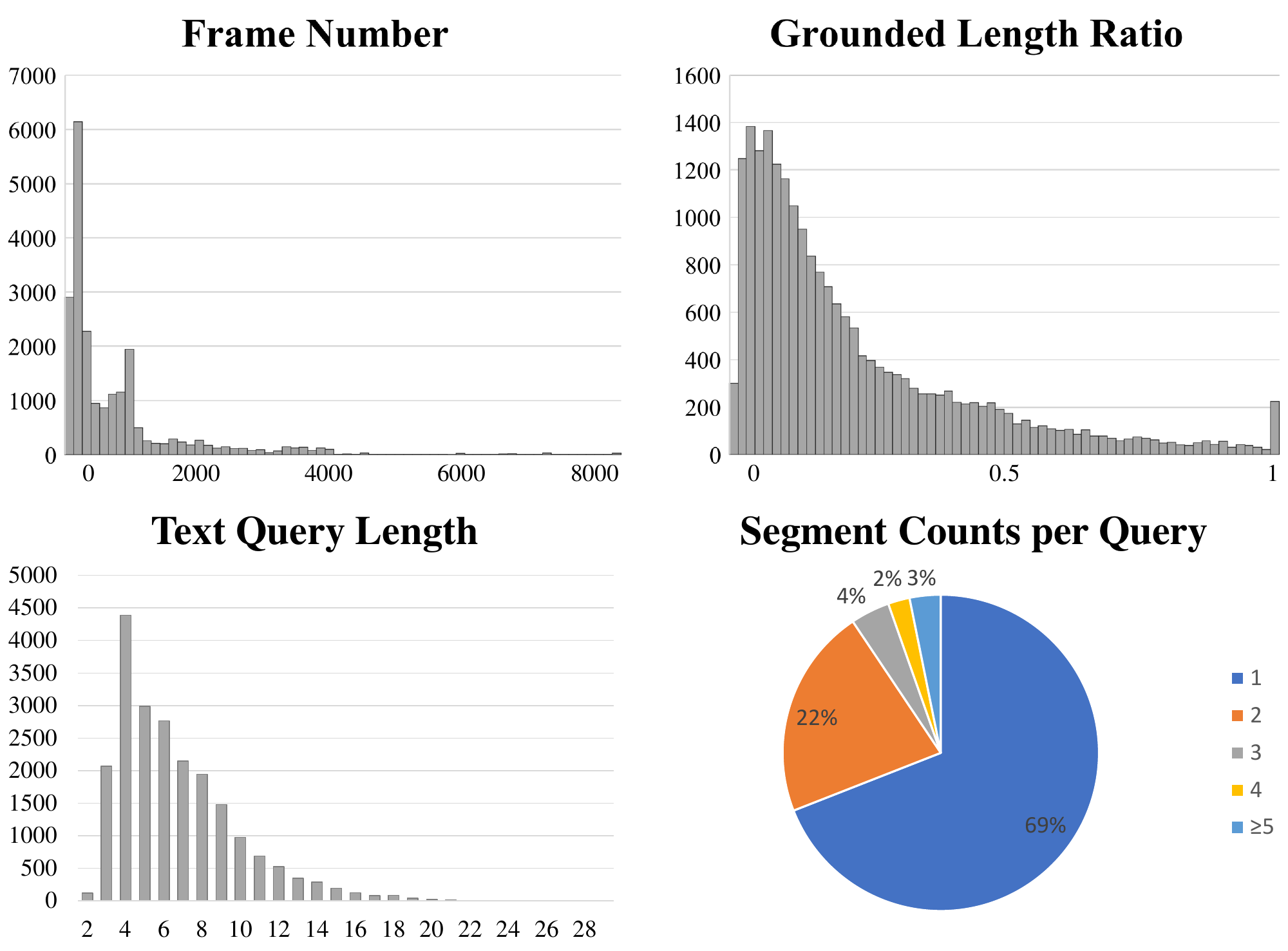}
\end{center}
   \caption{\small Dataset statistics of BABEL-Grounding. `Frame Number' refers to the length of motion sequences. `Text Query Length' denotes the length of textual annotations in the data. `Grounded Length Ratio' indicates the ratio of the length of temporal segments corresponding to each text query to the total length of the sequence. `Segment Counts per Query' refers to the number of temporal segments corresponding to each text query. }
\label{fig:dataset}
\end{figure}

\section{BABEL-Grounding Dataset}
The task of THMG involves determining the start and end timestamps of all the segments in the motion sequence that align with the given textual description. Our objective is to construct a dataset where each text query depicting specific human actions can be mapped to one or more temporal segments within the motion sequence. To this end, we construct the BABEL-Grounding dataset based on BABEL~\cite{punnakkal2021babel}. BABEL-Grounding provides detailed textual descriptions of human actions alongside their corresponding temporal segments in untrimmed motion sequences, with a total of 5,339 sequences with 21,307 text-segments annotations. Each sequence averages 743 frames, with the ground-truth temporal segments having an average frame count of 112, indicating their sparse distribution within the lengthy motion sequences. Figure~\ref{fig:dataset} illustrates the dataset statistics of BABEL-Grounding. As depicted, the BABEL-Grounding dataset contains comprehensive textual descriptions for motion sequences of diverse lengths. Each text query may correspond to multiple temporal segments, and these segments are sparsely distributed throughout the entire motion sequence, inherently posing a challenge for motion grounding task. 

\begin{figure}[t]
\begin{center}
\includegraphics[width=1.0\linewidth]{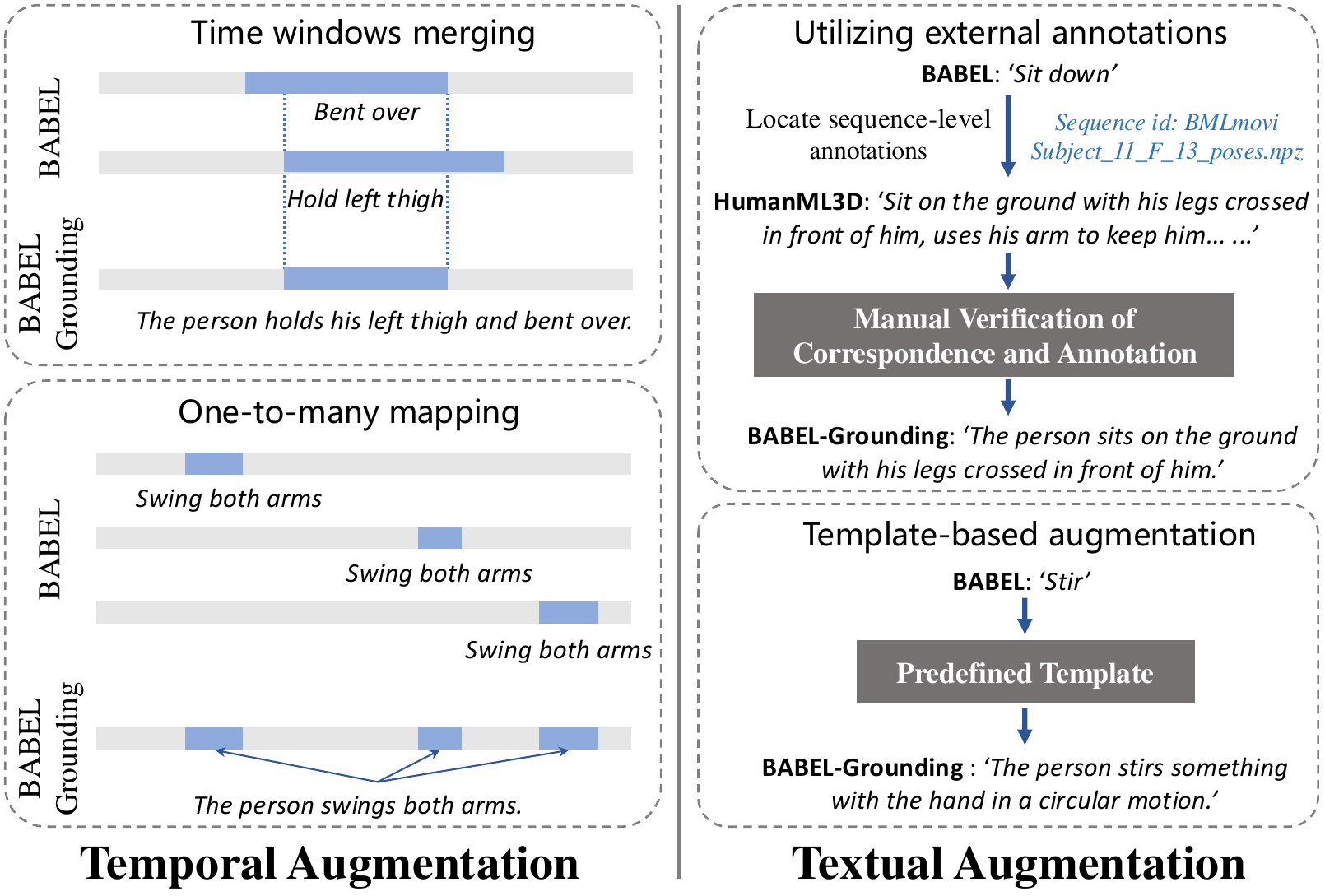}
\end{center}
   \caption{\small An illustration of the data augmentation pipeline, highlighting the differences between the original BABEL annotations and the BABEL-Grounding annotations.}
\label{fig:dataeval}
\end{figure}

Below, we provide a concise overview of the construction process for BABEL-Grounding annotations. The original BABEL dataset provides dense temporal annotations by labeling each temporal segment with corresponding actions. However, its textual annotations consist of simple categorical phrases rather than detailed and comprehensive sentences, and its data structure is not directly applicable to the THMG task. Therefore, a set of manually-crafted rules has been employed to augment the dataset. Figure~\ref{fig:dataeval} provides a schematic illustration of the data annotation pipeline and highlights the distinction between BABEL and BABEL-Grounding datasets. Details will be elaborated below.

\subsection{Textual augmentation}
The quality of textual annotations in BABEL is greatly limited by items that consist of only simple words or phrases, such as `place', `turn', and `step'. These ambiguous and meaningless items make up a considerable proportion of the data and fail to provide detailed descriptions of human body movement involved in the motion. To address this issue, two approaches have been applied:

\paragraph{Utilizing external annotations} BABEL is built upon the AMASS motion capture database~\cite{mahmood2019amass}, while HumanML3D~\cite{guo2022generating} offers detailed textual annotations at the sequence level for the AMASS database. For each entry in the BABEL dataset with overly simplistic annotations, its corresponding entry in HumanML3D is located by the sequence ID. When the sequence-level HumanML3D annotations contain the phrase from the BABEL annotations, human annotators will manually verify their correspondence and supplement the BABEL annotations with detailed textual descriptions from HumanML3D.

\paragraph{Template-based augmentation} As HumanML3D's sequence-level annotations only partially cover the items in BABEL, many low-quality annotations remain unaddressed, particularly those with only one-word labels. To address this gap, we've manually crafted templates to enhance them. For example, `take' is expanded to `take something with his hands', `stir' becomes `stir something with his hands in a circular motion', and `place' becomes `place the object at a specific location'. This substitution process enriches the overly simplistic annotations, rendering them more comprehensible for the model. The final textual annotations are further refined by \textit{GPT-3.5-turbo} to make them more fluent, complete and diverse.

\subsection{Temporal Augmentation}
\paragraph{Time windows merging} In the original BABEL dataset, temporal segments of multiple annotated items may overlap with each other. To further improve the quality of the textual annotations, we design a time windows merging rule, which merges annotated segments that have a significant overlap. To be specific, if the overlapped portion of two segments counts for more than a certain ratio (which is empirically set to 0.8) of one of them, the textual annotations of them are then merged to describe the motion within the overlapped part. This leads to more detailed texts which comprehensively depict the human motion.

\paragraph{one-to-many mapping} BABEL dataset provides annotations for each motion that occur in a sequence. However, in the THMG task, one query may correspond to multiple temporal segments in the sequence. To implement this feature, a one-to-many mapping from text to segments is established by merging the items with the same textual annotations. 

\subsection{Annotation Quality}
\subsubsection{Automatic Evaluation}

\begin{figure}[t]
\begin{center}
\includegraphics[width=1.0\linewidth]{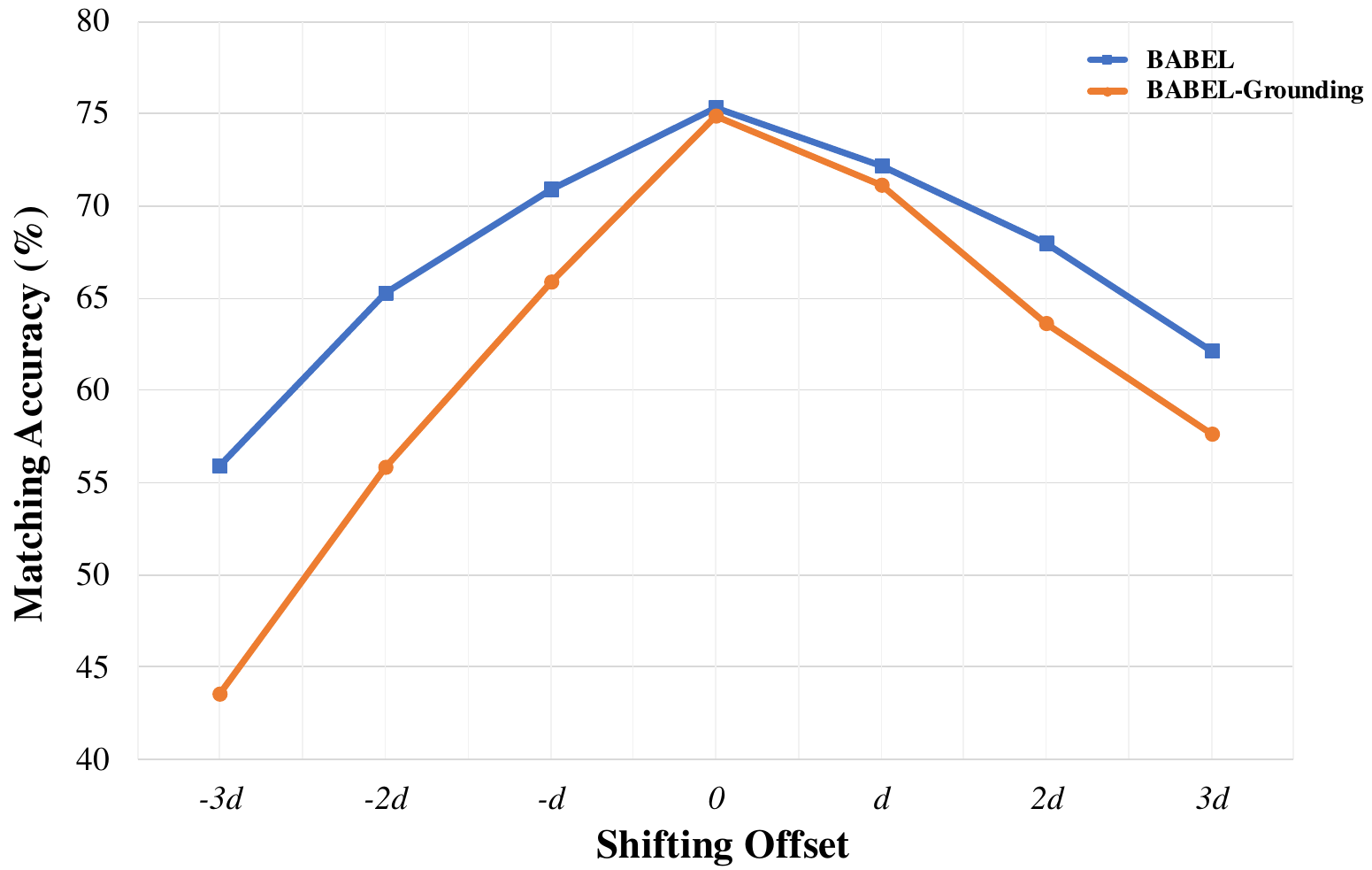}
\end{center}
   \caption{\small Evaluation results of annotation quality. $d = \frac{l}{3}$ measures the shifting offset relative to the time window length $l$. BABEL-Grounding achieves comparable matching accuracy to BABEL at zero offset, while demonstrating greater sensitivity to temporal shifts due to its richer information in text annotations.}
\label{fig:shifteval}
\end{figure}

To comprehensively evaluate the annotation quality of the proposed BABEL-Grounding dataset, we harness the power of Multimodal Large Language Model (MLLM) and design a protocol to assess both textual and temporal annotations. We select the open-source \textit{Qwen2.5-VL-72B-Instruct}~\cite{Qwen25VL} for its superior video understanding capabilities, running at BF16 precision on two NVIDIA A100 GPUs.

For each data item, we first obtain the rendered 3D mesh video of the motion and segment the video according to the annotated time window to produce a clip. Next, we feed the clip along with its corresponding textual description into the MLLM, which outputs a `yes' or `no' judgment on whether the motion in the clip matches the text. We calculate the proportion of `yes' responses as a matching accuracy indicator. 

To evaluate the quality of our temporal annotations, we design a method based on time window shifting. Specifically, the annotated time windows are shifted by fractions of the window length $l$, \textit{i.e.} ($\pm\frac{l}{3}$, $\pm\frac{2l}{3}$, $\pm l$), and then the matching accuracy with the text are computed using the same approach described above. The resulting accuracy curve, spanning from $-l$ to $+l$, is plotted in Figure~\ref{fig:shifteval}, with the curve for the original BABEL also presented for comparison. As expected, maximum alignment is observed at zero offset, with accuracy steadily declining as the displacement increases. This indicates that shifting our annotated time window causes a mismatch between the text and the motion within the window, validating the precision of our temporal annotations.

Furthermore, compared to the fully manual annotations from the BABEL dataset, our automatic augmentation approach achieves annotation quality that is nearly equivalent to human-level standards (75.3\% for BABEL and 74.9\% for BABEL-Grounding at zero offset). We also note that as the shift distance increases, the accuracy gap between BABEL and BABEL-Grounding widens. This disparity arises because BABEL provides short, concise and sometimes ambiguous phrases, while our dataset uses detailed, complete sentences with richer information. Consequently, our text-motion alignment is more sensitive to temporal mismatches, as reflected in the steeper decline in matching accuracy when the time windows are shifted.

\begin{figure}[t] 
  \centering
  \includegraphics[width=\linewidth]{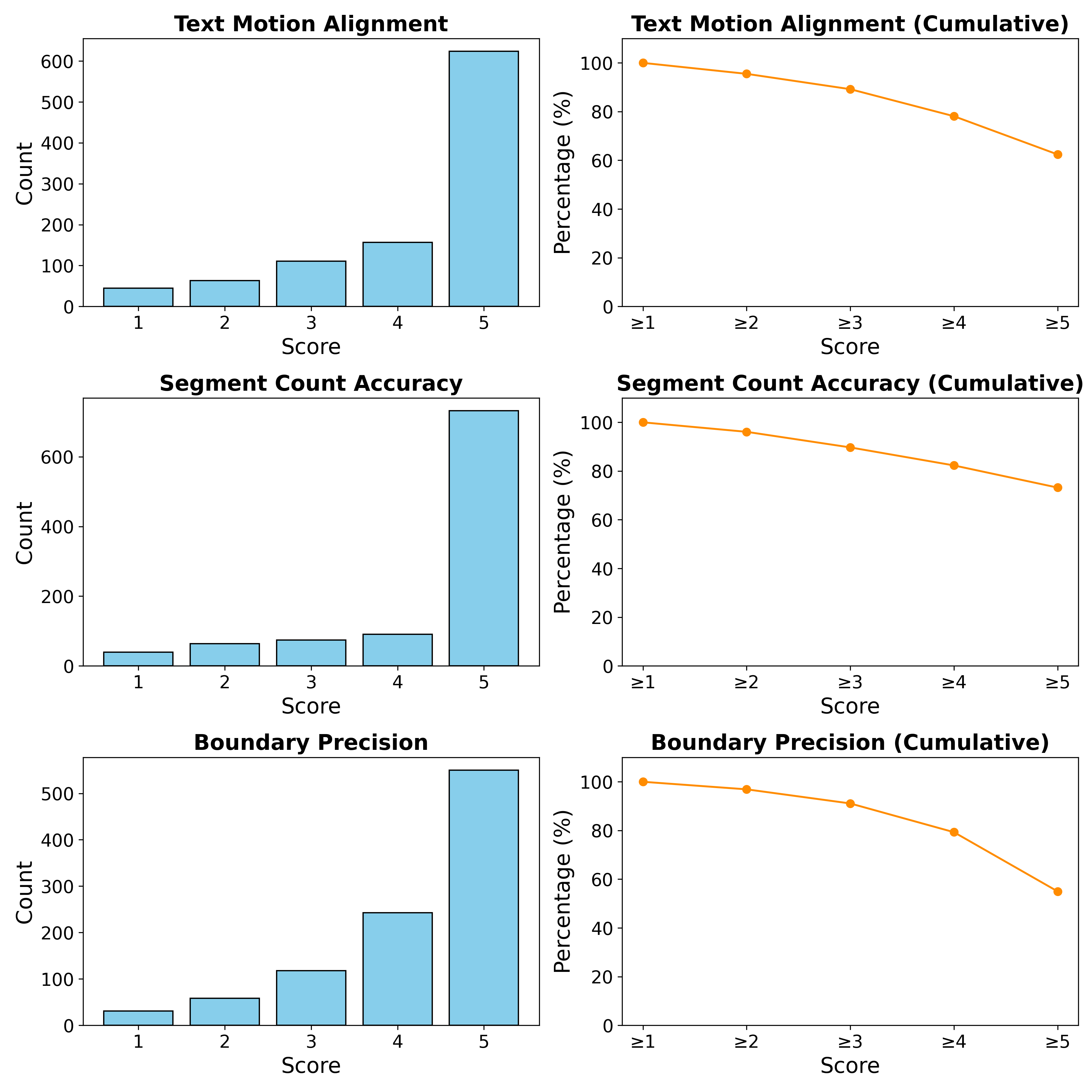}
  \caption{\textbf{Results of dataset evaluation.} The left side presents the frequency of scores ranging from 1 to 5, while the right side presents the cumulative percentages.}
  \label{fig:results}
\end{figure} 

\begin{table}[t]
    \centering
    \small
    \resizebox{1.0\linewidth}{!}
  {
    \begin{tabular}{lcccc}
        \toprule
        \textbf{Score Statistics} & \textbf{Mean} & \textbf{Std} & \textbf{Median} & \textbf{Truncated mean} \\
        \midrule
        Text-motion Alignment & 4.252 & 1.153 & 5.0 & 4.288  \\
        Segment Count Accuracy & 4.413 & 1.115 & 5.0 & 4.514 \\
        Boundary Precision  & 4.223 & 1.063 & 5.0 & 4.270  \\
        \bottomrule
    \end{tabular}
    }
    \caption{Score statistics of dataset evaluation, including the mean, variance, median, and the truncated mean.}
    \label{tab:statistics}
\end{table}

\begin{figure*}[t] 
  \centering
  \includegraphics[width=0.8\linewidth]{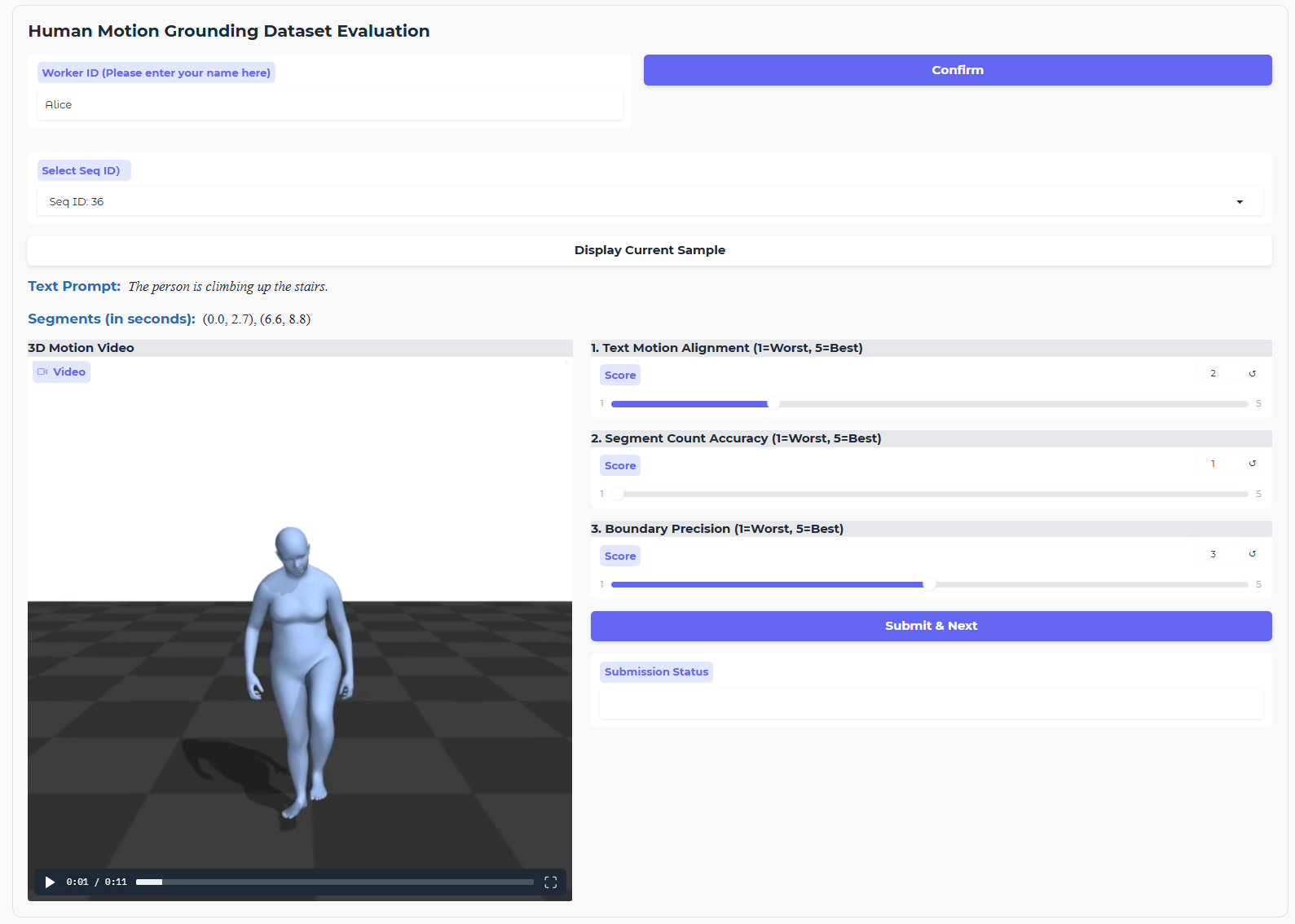}
  \caption{\textbf{Interactive Gradio interface of the dataset evaluation system.} Participants are presented with a video of a motion sequence together with its corresponding textual description and temporal segment annotations, and are required to provide rating scores for three questions.}
  \label{fig:interface}
\end{figure*}

\subsubsection{Manual Evaluation}
we conducted a thorough manual evaluation to assess the quality of the dataset annotations. To be specific, we invited 25 participants and randomly sampled 1,000 instances from the dataset. Each participant was asked to evaluate 40 samples, rating the annotations with scores from 1 to 5 (where 1 is the worst and 5 is the best) across three aspects:
\begin{itemize}
    \item Text-Motion Alignment: semantic consistency between the motion and the textual description.
    \item Segment Count Accuracy: correctness of the number of grounded temporal segments.
    \item Boundary Precision: accuracy of the start and end time boundaries of each segment.
\end{itemize}

The evaluation interface was implemented by Gradio (see Fig.~\ref{fig:interface}). Fig.~\ref{fig:results} presents the distribution of ratings across the three aspects. As shown, approximately 80\% of the samples received scores $\ge 4$, and about 90\% of the samples received scores $\ge 3$. Tab.~\ref{tab:statistics} summarizes statistical measures including mean, variance, and median scores for each aspect. To further account for the potential influence of extreme raters, we additionally report a truncated mean by excluding the top-3 and bottom-3 participants in terms of their overall average scores. In summary, both the automatic MLLM-based evaluation and the above manual evaluation consistently confirm the quality of the textual and temporal annotations in our BABEL-Grounding dataset.

\begin{figure*}[t]
\begin{center}
\includegraphics[width=0.80\linewidth]{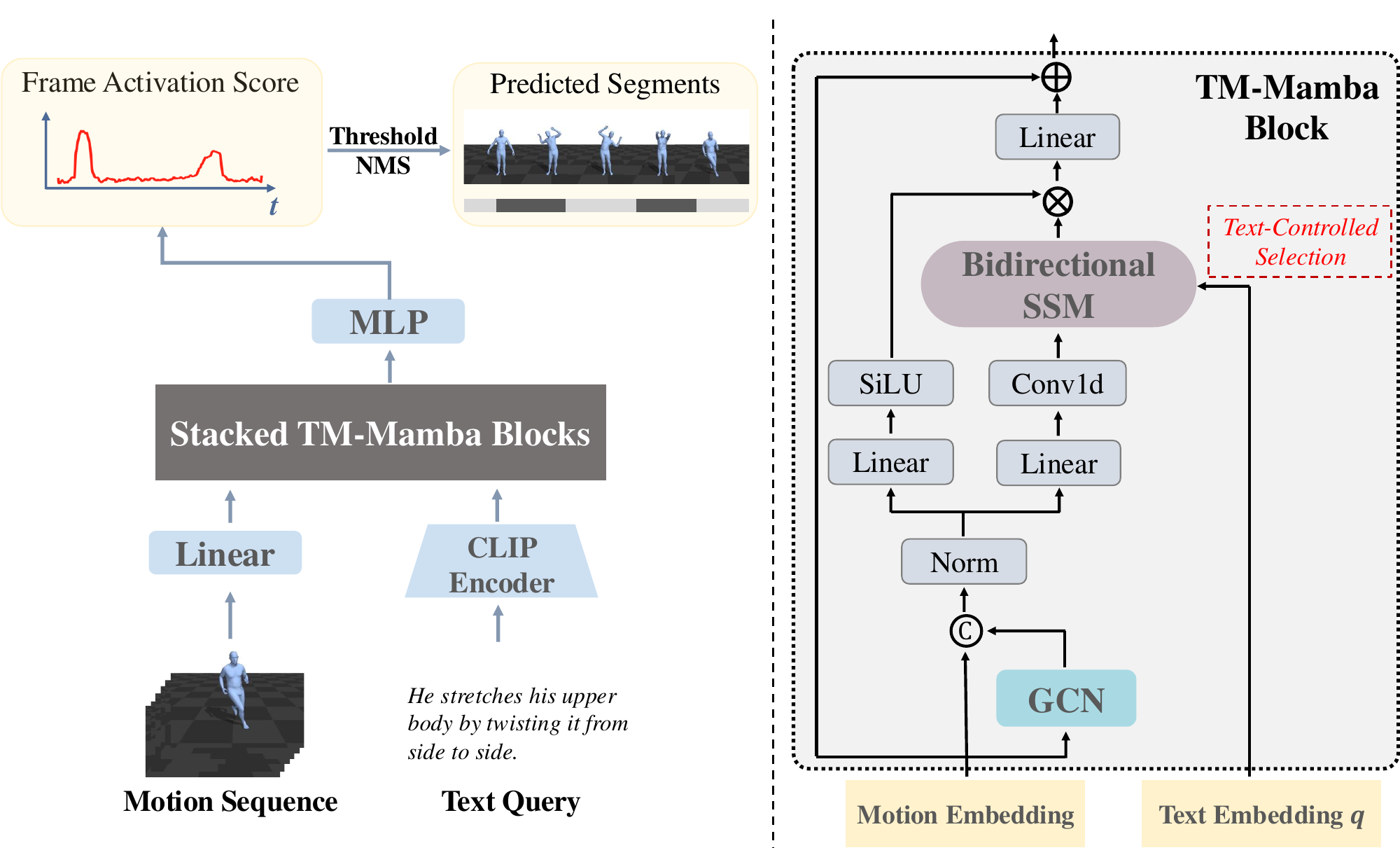}
\end{center}
   \caption{\small Left: overall architecture of our proposed model. Right: Illustration of TM-Mamba block. `Bidirectional SSM' refers to the text-controlled selection mechanism demonstrated in Algorithm~\ref{alg:our} with bidirectional modeling. Text embedding \textit{q} denotes the CLIP embedding of the input text query.}
\label{fig:overall}
\end{figure*}

\begin{figure*}[h]
\begin{minipage}{0.49\textwidth}
\centering
\begin{algorithm}[H]
\caption{Selection Mechanism (Mamba)}
\label{alg:mamba}
\begin{algorithmic}[1]

\Require{Input sequence $\mathbf{X} \in \mathbb{R}^{\mathtt{V} \times \mathtt{L} \times \mathtt{D}}$}
\Ensure{Output sequence $\mathbf{Y} \in \mathbb{R}^{\mathtt{V} \times \mathtt{L} \times \mathtt{D}}$}

\State $\mathbf{B}$ : $(\mathtt{V}, \mathtt{L}, \mathtt{N}) \leftarrow \mathrm{Linear}_{\mathbf{B}}(\mathbf{X})$
\State $\mathbf{C}$ : $(\mathtt{V}, \mathtt{L}, \mathtt{N}) \leftarrow \mathrm{Linear}_{\mathbf{C}}(\mathbf{X})$
\State $\mathbf{\Delta}$ : $(\mathtt{V}, \mathtt{L}, \mathtt{D}) \leftarrow \log(1 + \exp(\mathrm{Linear}_{\mathbf{\Delta}}(\mathbf{X}) + \mathrm{Parameter}_{\mathbf{\Delta}}))$
\State $\overline{\mathbf{A}}, \overline{\mathbf{B}}$ : $(\mathtt{V}, \mathtt{L}, \mathtt{D}, \mathtt{N}) \leftarrow \mathrm{discretize}(\mathbf{\Delta}, \mathrm{Parameter}_{\mathbf{A}}, \mathbf{B})$
\State $\mathbf{Y} \leftarrow \mathrm{SSM} (\overline{\mathbf{A}}, \overline{\mathbf{B}}, \mathbf{C}) (\mathbf{X})$
\State \Return $\mathbf{Y}$
\end{algorithmic}
\end{algorithm}
\end{minipage}\hfill
\begin{minipage}{0.49\textwidth}
\centering
\begin{algorithm}[H]
\caption{Text-Controlled Selection Mechanism}
\label{alg:our}
\begin{algorithmic}[1]
\Require{Input sequence $\mathbf{X} \in \mathbb{R}^{\mathtt{V} \times \mathtt{L} \times \mathtt{D}}$; Text embedding $q \in \mathbb{R}^{D}$}
\Ensure{Output sequence $\mathbf{Y} \in \mathbb{R}^{\mathtt{V} \times \mathtt{L} \times \mathtt{D}}$}

\State $\mathbf{B}$ : $(\mathtt{V}, \mathtt{L}, \mathtt{N}) \leftarrow \mathrm{Linear}_{\mathbf{B}}(\mathbf{X}, \textcolor{shapecolor}{q})$
\State $\mathbf{C}$ : $(\mathtt{V}, \mathtt{L}, \mathtt{N}) \leftarrow \mathrm{Linear}_{\mathbf{C}}(\mathbf{X}, \textcolor{shapecolor}{q})$
\State $\mathbf{\Delta}$ : $(\mathtt{V}, \mathtt{L}, \mathtt{D}) \leftarrow \log(1 + \exp(\mathrm{Linear}_{\mathbf{\Delta}}(\mathbf{X}, \textcolor{shapecolor}{q}) + \mathrm{Parameter}_{\mathbf{\Delta}}))$
\State $\overline{\mathbf{A}}, \overline{\mathbf{B}}$ : $(\mathtt{V}, \mathtt{L}, \mathtt{D}, \mathtt{N}) \leftarrow \mathrm{discretize}(\mathbf{\Delta}, \mathrm{Parameter}_{\mathbf{A}}, \mathbf{B})$
\State $\mathbf{Y} \leftarrow \mathrm{SSM} (\overline{\mathbf{A}}, \overline{\mathbf{B}}, \mathbf{C}) (\mathbf{X})$ \label{ssm}
\State \Return $\mathbf{Y}$
\end{algorithmic}
\end{algorithm}
\end{minipage}
\end{figure*}

\section{Method}
\subsection{Preliminaries on Mamba}
In this section, we present a brief review of State Space Models (SSM) and Mamba~\cite{gu2023mamba}. SSM is a series of sequential models that maps the input sequence $x(t) \in \mathbb{R}$ to the output sequence $y(t) \in \mathbb{R}$ through a hidden state $h(t) \in \mathbb{R}^N$, which can be depicted as a linear ODE:
\begin{equation}
\begin{aligned}
\label{eq:lti}
h'(t) &= \mathbf{A}h(t) + \mathbf{B}x(t), \\
y(t) &= \mathbf{C}h(t).
\end{aligned}
\end{equation}
where $\mathbf{A} \in \mathbb{R}^{N \times N}, \mathbf{B} \in \mathbb{R}^{N \times 1}, \mathbf{C} \in \mathbb{R}^{1 \times N}$ are the evolution and projection parameters. This continuous ODE can be discretized using a timescale parameter $\mathbf{\Delta}$ following the zero-order hold (ZOH) rule:
\begin{equation}
\begin{aligned}
\label{eq:zoh}
\mathbf{\overline{A}} &= \exp{(\mathbf{\Delta}\mathbf{A})}, \\
\mathbf{\overline{B}} &= (\mathbf{\Delta} \mathbf{A})^{-1}(\exp{(\mathbf{\Delta} \mathbf{A})} - \mathbf{I}) \cdot \mathbf{\Delta} \mathbf{B}.
\end{aligned}
\end{equation}
The discretized form of the aforementioned formulation can be calculated using linear recurrence:
\begin{equation}
\begin{aligned}
\label{eq:discrete_lti}
h_t &= \mathbf{\overline{A}}h_{t-1} + \mathbf{\overline{B}}x_{t}, \\
y_t &= \mathbf{C}h_t.
\end{aligned}
\end{equation}

However, linear recurrence requires unfolding in time and is unable to be parallelized. S4~\cite{gu2021efficiently} ensures Linear Time Invariance (LTI) by assuming that $\mathbf{A}, \mathbf{B}, \mathbf{C}, \mathbf{\Delta}$ remain static, allowing for the implementation using global convolution as $\mathbf{y} = \mathbf{x} * \mathbf{\overline{K}}$, where
\begin{equation}
\begin{aligned}
\label{eq:conv}
\mathbf{\overline{K}} &= (\mathbf{C}\mathbf{\overline{B}}, \mathbf{C}\mathbf{\overline{A}}\mathbf{\overline{B}}, \dots, \mathbf{C}\mathbf{\overline{A}}^{L-1}\mathbf{\overline{B}}), 
\end{aligned}
\end{equation}
and $L$ denotes the length of the input sequence, and $\overline{\mathbf{K}} \in \mathbb{R}^{L}$ represents a structured convolutional kernel.
On the other hand, Mamba~\cite{gu2023mamba} introduces an input-dependent selection mechanism by making $\mathbf{\overline{A}}, \mathbf{\overline{B}}, \mathbf{C}, \mathbf{\Delta}$ become functions of $x_t$. Such formulation can be efficiently computed via the proposed parallel scan algorithm.

\subsection{Text-Controlled Selection Mechanism}
Mamba emphasizes the crucial role of selectivity in constructing sequence models. When dealing with very long sequences, it becomes impractical to memorize all the information within the compressed state vector. Therefore, it's essential to design a selection mechanism that controls how information propagates or interacts along the sequence dimension. Mamba addresses this challenge by employing a context-aware parameterization of state transition matrices. This enables the model to focus on or filter out information based on the current input data. 

However, in the context of the THMG task, the model needs to dynamically select relevant global information from the sequence based on the text query to achieve better grounding performance. Existing multimodal Mamba methods~\cite{qiao2024vl, zhao2024cobra} use simple concatenation to merge textual and visual features as input for Mamba, but the key selection mechanism governing the information flow remains the same. To overcome this limitation, we propose a text-controlled selection mechanism that allows the selection process to depend on both the motion input and the text query. Given the input text query, we first utilize CLIP~\cite{radford2021learning} to acquire the text embedding $q$. As illustrated in Algorithm~\ref{alg:mamba} and~\ref{alg:our}, Mamba parameterizes $\mathbf{\overline{A}}, \mathbf{\overline{B}}, \mathbf{C}, \mathbf{\Delta}$ as functions of the input, whereas in text-controlled selection, these parameters become functions of input sequence as well as text embedding $q$.

The algorithm essentially resembles a text-based gating mechanism, dynamically controlling the flow of information based on textual queries. Theorem 1 of~\cite{gu2023mamba} implies that Algorithm~\ref{alg:our} exhibits similarities to gated RNN under certain conditions:
\begin{lemma}
\label{lemma:gate}
When $N = 1, \mathbf{A} = -1, \mathbf{B}=1$, the text-controlled selection mechanism takes the form of $g_t = \sigma(Linear_{\Delta}(X, q))$ and $h_t = (1-g_t) h_{t-1} + g_t h_t$, where X denotes input sequence and q denotes query embedding.
\end{lemma}
Lemma~\ref{lemma:gate} indicates that the text-controlled selective SSM bears resemblance to a gated RNN, wherein the gate $g_k$ relies on both motion input and text query, enabling the text query to control the information flow during propagation. Given that the recurrence process in line 5 of Algorithm~\ref{alg:our} remains unchanged, the resultant text-controlled SSM can still be efficiently computed using the parallel scan algorithm outlined in~\cite{gu2023mamba}. The new algorithm can be implemented by modifying the forward and gradient backward functions of the original Mamba, which enables end-to-end joint training of SSM and the language backbone.

\begin{table*}[t]
\caption{Ablation studies on BABEL-Grounding dataset. The best results are in bold. 
} 
	{\small
		\centering
			\resizebox{0.85\textwidth}{!}{
				\begin{tabular}{l|c|c|cccccccc}
					\toprule
					\multirow{2}{*}{\textbf{Methods}} & \multirow{2}{*}{\textbf{Text-Controlled}}& \multirow{2}{*}{\textbf{Relational}} &\multicolumn{8}{c}{\textbf{mAP@IoU (\%)}} \\
					& &  & 0.1   & 0.2 & 0.3 & 0.4 & 0.5 & 0.6 & 0.7 & Average\\ \midrule \midrule
\multirow{4}*{Unidirectional}  & & &  29.8& 27.2 & 24.9 & 22.8& 20.5& 17.8 & 14.4 & 22.5\\

 & \checkmark & &41.5 & 37.9 & 34.4 & 30.7 & 27.7 & 23.7 & 18.7 & 30.7 \\
 &  &\checkmark & 40.5& 37.1 & 34.0 & 30.6 & 27.7 & 23.9 & 19.6 & 30.5 \\
 & \checkmark &\checkmark & 42.5& 38.5 & 34.9 & 31.5& 28.3 & 23.9& 19.2 &  31.3\\
 \midrule
  \multirow{4}*{Bidirectional}  & &  & 38.3 & 35.0 & 32.0  & 29.2 & 26.8 & 23.4 & 19.0 & 29.1 \\
   & \checkmark &  & 51.2 & 48.0 & 44.1 & 40.1 & 36.0 & 30.5 & 24.7 &  39.2\\
 &  &\checkmark & 46.8&  43.6 & 40.0 &  35.5& 32.0 & 27.1 & 21.2 & 35.2 \\
 & \checkmark &\checkmark & \textbf{53.9} &  \textbf{50.5} & \textbf{46.7}& \textbf{42.8} & \textbf{38.4} & \textbf{32.6} & \textbf{26.0} &  \textbf{41.6}\\
  \toprule
				\end{tabular}
			}

		\label{tab:ablation}
	}

\end{table*}

\subsection{Text-Controlled Motion Mamba}
Text-Controlled Selective SSMs enjoy linear computational complexity and memory consumption, making them suitable for extracting the global context of very long sequences. This makes it a natural choice for temporal modeling in motion grounding task. However, the human skeleton inherently possesses a latent graph structure, constituting a multivariate time series. Mamba operates on univariate sequences, thus overlooking the interaction of human joints.

In this study, we enhance Mamba through the incorporation of topology awareness, achieved by integrating relational embeddings to convey information regarding neighboring nodes. 
The input skeleton sequence is first embedded by a linear layer, resulting in an embedding sequence with shape $\mathbf{X} \in \mathbb{R}^{V \times L \times D}$, where $V$ denotes the number of joints in the human skeleton, $L$ denotes the sequence length, and $D$ denotes the feature dimension.  
The graph-structured representation of human skeleton has been extensively validated in prior research, with Graph Convolutional Networks (GCNs) demonstrating exceptional capability in modeling inter-joint relational patterns~\cite{chen2021channel, chi2022infogcn, liu2020disentangling, shi2019two, yan2018spatial, hao2021hypergraph, cheng2021extremely}. Building upon these advancements, we compute the relational embedding as $\mathbf{R} = f(\mathbf{X}) \in \mathbb{R}^{V \times L \times D}$, which encapsulates the graph message at each time step. Here, $f$ is implemented as the Adaptive Graph Convolutional Network (AGCN) proposed in~\cite{shi2019two}, allowing the skeleton graph topology to be learned adaptively. Subsequently, the relational embedding $\mathbf{R}$ is concatenated with the motion features and fed into the text-controlled selective SSM in Algorithm~\ref{alg:our}, to jointly capture the global temporal information of each joint alongside its topological context.

The overall architecture of the Text-Controlled Motion Mamba is depicted in Figure~\ref{fig:overall}. Unlike the vanilla Mamba block, which adopts unidirectional causal modeling, the task of THMG demands the global context of the entire sequence. To address this issue, a bidirectional non-causal structure, as proposed in Vision Mamba~\cite{zhu2024vision}, is employed. The input goes through a stack of TM-Mamba blocks, producing the output $\mathbf{Y} \in \mathbb{R}^{V \times L \times D}$.  After mean pooling along the $V$ dimension, $\mathbf{Y}$ is subsequently forwarded to an MLP layer. This yields the frame activation score $s_t (t=1, 2, \ldots, T)$ for each frame, indicating the likelihood of its inclusion in the retrieved temporal segments. The entire framework can be supervised using a simple cross-entropy loss.
\begin{equation}
    \mathcal{L}_{ce} = -\frac{1}{T} \sum_{t}^T (y_t \log{s_t} + (1 - y_t) \log{(1 - s_t)} ).
\end{equation}
where $y_t$ denotes the ground-truth label indicating whether frame $t$ lies within the segments retrieved by the text query. The proposed method achieves effective global context extraction, query-based information selection, and topology modeling within a unified yet simple framework. Compared to Transformer-based methods, our approach eliminates the necessity of computing self-attention over the entire sequence along the temporal dimension. This eliminates the need for quadratic memory, rendering it feasible for processing very long sequences.

\section{Experiments}
\subsection{Implementation and Evaluation Details}
The motion data pre-processing procedure adheres to the methodology outlined in BABEL~\cite{punnakkal2021babel}. The maximum length of the motion sequence is constrained to 2000. The same data split for training and evaluation as BABEL is adopted. CLIP~\cite{radford2021learning} is adopted as text encoder to acquire the query embedding $q$ in Algorithm~\ref{alg:our}. To ensure fair comparison, we utilized the same text encoder throughout the experiments, namely \textit{clip-vit-base-patch32}, with its parameters being jointly tuned together with the entire model during training. The feature dimension $D$ in Algorithm~\ref{alg:our} for both the input motion sequence and text embedding is set to 256. The batch size and base learning rate are configured as $4$ and $5 \times 10^{-4}$ respectively, with the learning rate for CLIP set to $5 \times 10^{-5}$. The optimizer is an AdamW~\cite{loshchilov2017decoupled} with a weight decay of $1 \times 10^{-4}$. The number of stacked Motion-Mamba blocks is empirically specified as $3$. Model training is conducted on a single Nvidia A40 GPU. 

During the inference stage, a series of thresholds are used to obtain the predicted temporal segments, following~\cite{caba2015activitynet}. Subsequently, non-maximum suppression is performed to remove overlapping segments. The evaluation of grounding performance is conducted using mean Average Precisions (mAPs) under different Intersection of Union (IoU) thresholds, namely $[0.1:0.7:0.1]$.

\subsection{Ablation Studies}
We conduct extensive ablative studies on BABEL-Grounding dataset to demonstrate the effect of each component in our proposed TM-Mamba model.

\noindent \textbf{Text-Controlled Selection Mechanism.} The text-controlled selection mechanism lies at the core of the TM-Mamba model. Removing this component directly from Algorithm~\ref{alg:our} for ablation causes the model incapable of perceiving text query information. To enable comparison, the ablative models follow the practice of~\cite{qiao2024vl, zhao2024cobra}, where text embeddings are concatenated with the sequence input. An MLP is then employed for feature fusion. As demonstrated in Table~\ref{tab:ablation}, the text-controlled selection mechanism accounts for a substantial performance gain. This underscores the significance of dynamically regulating the propagation of information based on textual queries.

\noindent \textbf{Unidirectional v.s. Bidirectional.} The Vanilla Mamba model adopts a unidirectional approach, whereby the model can only access the sequence history while processing the current step. This unidirectional approach proves advantageous for tasks such as language modeling. However, in the context of the THMG task, the model requires a comprehensive understanding of the global context of the entire sequence, demanding a non-causal bidirectional structure. The results presented in Table~\ref{tab:ablation} substantiate this claim, showing that the bidirectional model significantly outperforms its unidirectional counterpart.

\noindent \textbf{Relational Embedding.} The importance of topology modeling in THMG task is also evaluated by removing the relational embedding from TM-Mamba. As shown, the inclusion of relational embeddings $\textbf{R}$ improves the performance of the model by effectively capturing the underlying graph structure within the skeletal data. 

\begin{table}[t]
\caption{Performance comparisons to baseline methods on BABEL-Grounding dataset. The best results are in bold. 
} 
	{\large
		\centering
			\resizebox{0.5\textwidth}{!}{
				\begin{tabular}{l|cccccccc}
					\toprule
					\multirow{2}{*}{\textbf{Methods}} & \multicolumn{8}{c}{\textbf{mAP@IoU (\%)}} \\
					& 0.1   & 0.2 & 0.3 & 0.4 & 0.5 & 0.6 & 0.7 & Avg\\ \midrule \midrule
     
S4D-LegS~\cite{gu2022parameterization}  & 29.1 & 26.2 & 23.0 & 19.6 & 15.9 & 12.0 & 8.9 & 19.2 \\
S4D-Lin~\cite{gu2022parameterization}  & 30.1& 26.9 & 23.8  & 20.5 & 16.6 & 12.8& 9.5 & 20.0\\
2s-AGCN~\cite{shi2019two}  &31.2  & 24.7 & 20.5& 17.1& 14.0 & 10.5 & 7.8& 18.0\\
InfoGCN~\cite{chi2022infogcn}  & 49.5 & 42.5  & 36.1  & 30.5 & 26.0  & 20.8 & 15.3  & 31.5 \\
MotionCLIP~\cite{tevet2022motionclip}  & 41.8 & 39.2 & 36.1 & 32.7 & 29.3  & 24.9 & 19.9  & 32.0 \\
MomentDETR~\cite{lei2021detecting}  & 51.1 & 46.0 & 39.1 & 32.9 & 26.8 & 20.5 & 13.6 & 32.9 \\
EaTR~\cite{jang2023knowing}  & 53.4  &  48.5  & 43.6  & 37.3  & 31.0  & 23.5  & 16.2  & 36.2  \\
STCAT~\cite{jin2022embracing}  & 47.1  & 44.2 & 40.7 & 37.1 & 33.3 & 28.9 & 23.6 & 36.4 \\

 \midrule
\textbf{TM-Mamba}  &\textbf{53.9} &  \textbf{50.5} & \textbf{46.7}& \textbf{42.8} & \textbf{38.4} & \textbf{32.6} & \textbf{26.0} &  \textbf{41.6}\\
  \toprule
				\end{tabular}
			}

		\label{tab:overall}
	}

\end{table}

\begin{figure}[t]
\begin{center}
\includegraphics[width=1.0\linewidth]{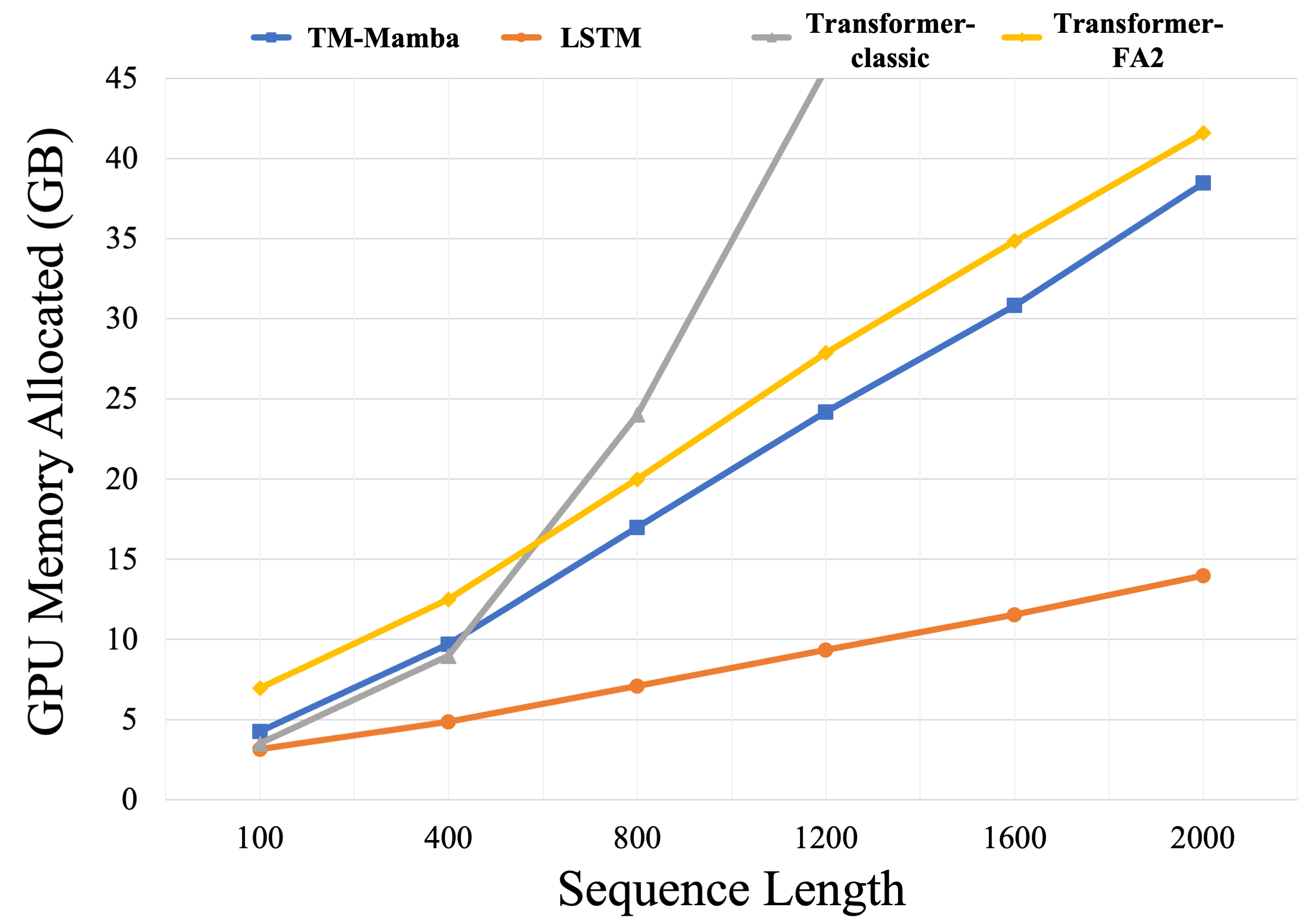}
\end{center}
   \caption{\small Comparison of memory consumption for TM-Mamba and its LSTM, classic Transformer and Flash-attention-2-based (FA2) Transformer  counterparts under varying motion sequence lengths. The classic Transformer runs out of GPU memory when the sequence length reaches 1200.}
\label{fig:memory}
\vspace{-1.0mm}
\end{figure}

\begin{table}[t]
\caption{Performance comparison of TM-Mamba and its temporal Transformer counterpart under different maximum sequence length.
} 
	{\large
		\centering
			\resizebox{0.5\textwidth}{!}{
				\begin{tabular}{cc|cccccccc}
					\toprule
					\multirow{2}{*}{\textbf{Length}} &  \multirow{2}{*}{\textbf{Model}} & \multicolumn{8}{c}{\textbf{mAP@IoU (\%)}} \\
					& & 0.1   & 0.2 & 0.3 & 0.4 & 0.5 & 0.6 & 0.7 & Avg\\ \midrule \midrule

\multirow{2}{*}{\textbf{300}} & TM-Mamba  &77.0 &73.7 &68.8 &63.0 &58.2 &50.2 &39.5 &61.5 \\
 & Transformer &72.9 &67.3 &60.9 &54.6 &48.4 &39.4 &30.2  & 53.4\\
\midrule
 \multirow{2}{*}{\textbf{500}} & TM-Mamba  &74.0 &70.2 &65.4 &59.7 &53.7 &45.7 &37.1 &58.0 \\
 & Transformer &67.3 &62.3 &56.7 &50.5 &44.7 &37.0 &28.2  &49.5 \\
\midrule
 \multirow{2}{*}{\textbf{1000}} & TM-Mamba  &58.0 &54.5 &50.5 &46.5 &41.4 &35.9 &28.4 &45.0 \\
 & Transformer &53.8 &48.4 &42.9 &37.7 &32.6 &26.8 &20.6  &37.5 \\
\midrule
 \multirow{2}{*}{\textbf{1500}} & TM-Mamba  &55.6 &51.9 &48.1 &44.2 &39.5 &33.8 &26.6 &42.8 \\
 & Transformer & \multicolumn{8}{c}{\small{\emph{-- Out of Memory --}}}\\
\midrule
 \multirow{2}{*}{\textbf{2000}} & TM-Mamba  &53.9 &50.5 &46.7 &42.8 &38.4 &32.6 &26.0 &41.6 \\
 & Transformer &\multicolumn{8}{c}{\small{\emph{-- Out of Memory --}}} \\

  \toprule
				\end{tabular}
			}

		\label{tab:length}
	}

\end{table}

\begin{figure}[t]
\begin{center}
\includegraphics[width=1.0\linewidth]{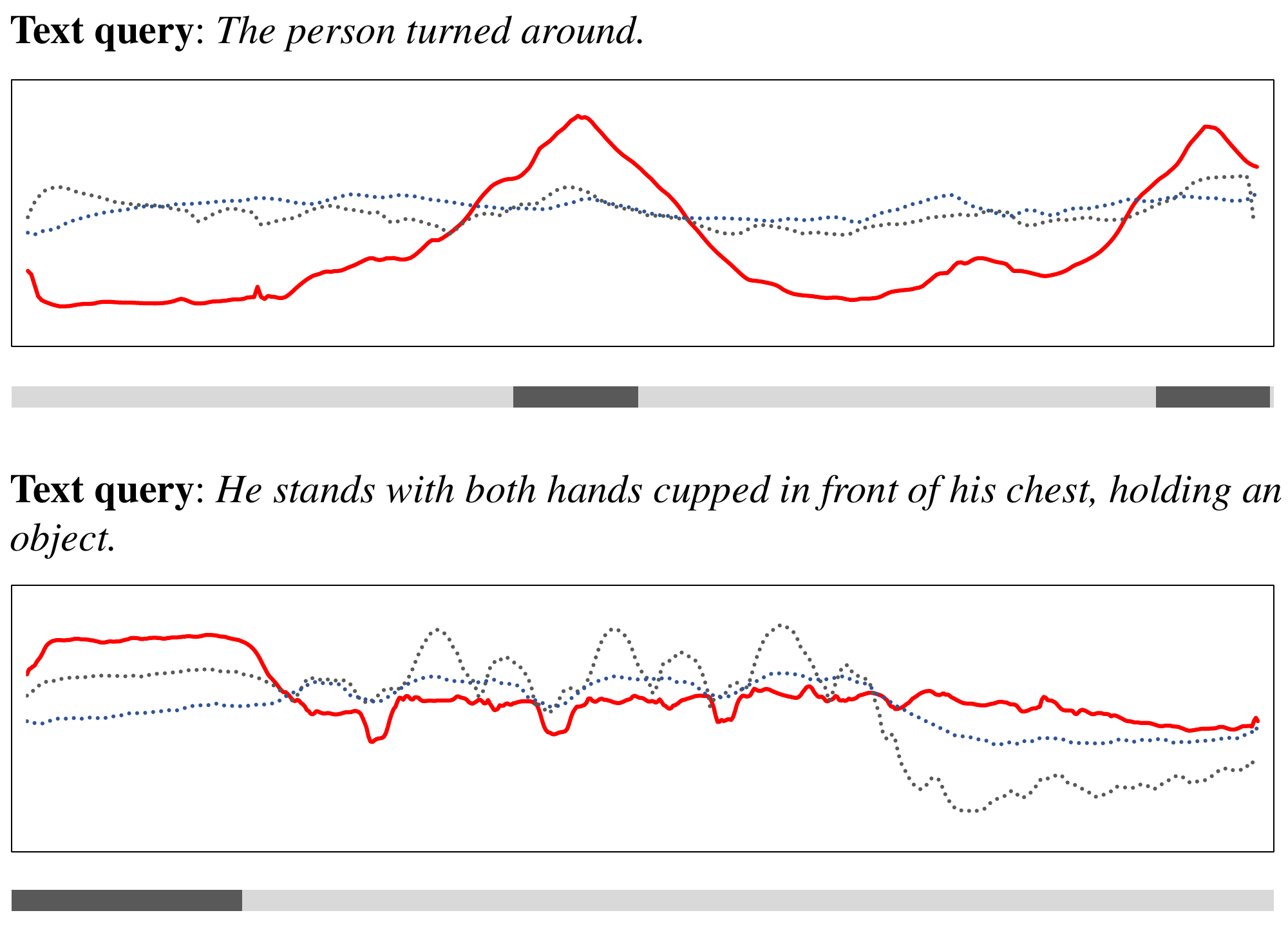}
\end{center}
   \caption{\small Visualizations of predicted frame activation score. The solid red line denotes the predicted score of our full model, while the dashed blue and gray lines denote the model without text control and relational embeddings, respectively. The gray bars below illustrate the ground-truth temporal segments. Best viewed in color.}
\label{fig:frame}
\end{figure}

\begin{figure*}[t]
\begin{center}
\includegraphics[width=0.8\linewidth]{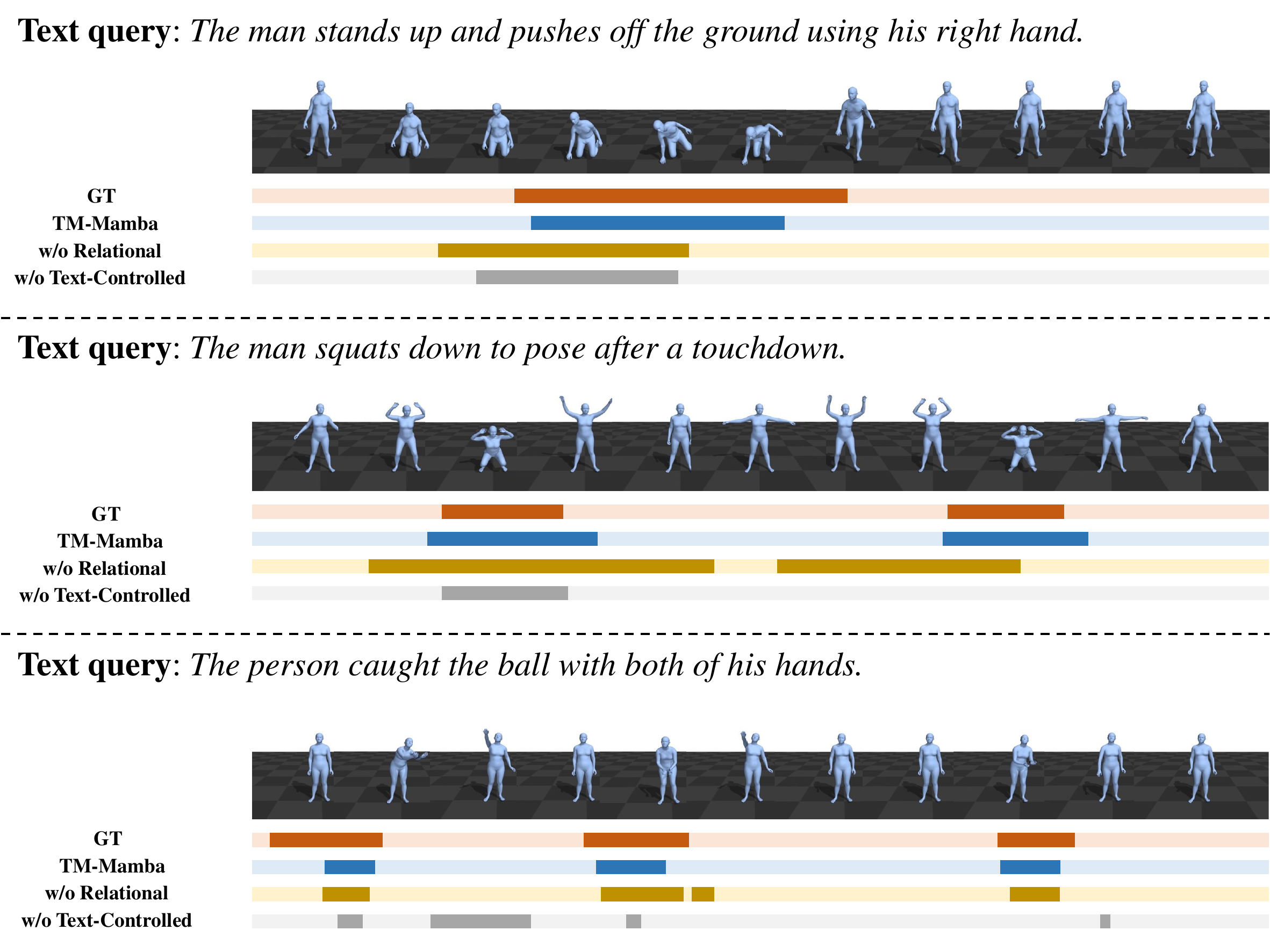}
\end{center}
   \caption{\small Visualizations of the grounding results of TM-Mamba compared to ablative models. `GT' denotes ground-truth temporal segments corresponding to the text query. Our TM-Mamba demonstrates superior performance in motion grounding, in terms of the number of retrieved segments and the temporal boundaries of each segment. Best viewed in color.}
\label{fig:visualization}
\end{figure*}

\subsection{Comparison with Baselines}
Given that the THMG task is novel and lacks existing works for comparison, several baseline methods are implemented for analysis. To begin, several prominent models are chosen from the video moment retrieval task: MomentDETR~\cite{lei2021detecting}, EaTR~\cite{jang2023knowing} and STCAT~\cite{jin2022embracing}. However, these models rely on global temporal Transformers, which require quadratic computational memory and are impractical for handling the lengthy sequences in THMG task. As a remedy, their global temporal Transformers are substituted with efficient GRUs which require only linear memory. Also, two baselines based on powerful human motion backbones are implemented using spatial-temporal graph convolutions, namely 2s-AGCN~\cite{shi2019two} and InfoGCN~\cite{chi2022infogcn}. Text embedding is incorporated through concatenation and MLP fusion to integrate query information for text-based grounding. We also implement a baseline where frame activation scores are computed using MotionCLIP~\cite{tevet2022motionclip}, a motion-language model that aligns CLIP text embeddings with motion features through cosine similarity loss. Finally, we implemented several baselines based on recent SSM-based models~\cite{gu2022parameterization} to demonstrate the advantages of TM-Mamba over other SSM methods. As illustrated in Table~\ref{tab:overall}, TM-Mamba outperforms the baseline models in terms of mAP at various IoU thresholds, thereby demonstrating its effectiveness.

\subsection{More Analysis}
\paragraph{Memory Consumption}
The memory usage of the Mamba-based models increases linearly with the length of the sequence, which allows them to effectively handle longer sequences. Figure~\ref{fig:memory} validates this by comparing the GPU memory consumption of our TM-Mamba model with its Transformer, LSTM and flash-attention~\cite{dao2022flashattention} counterparts when processing sequences of varying lengths. As depicted in the figure, the memory cost of the Transformer-based model increases quadratically with sequence length, causing it to quickly run out of memory. On the other hand, TM-Mamba, LSTM and flash-attention-based efficient transformer exhibit linear memory consumption, making them applicable for processing long motion sequences when extracting global context information. The slope of TM-Mamba is larger than its LSTM counterpart due to larger hidden states.

\paragraph{Comparison with Transformer}
Table~\ref{tab:length} presents a comparison between TM-Mamba and its temporal Transformer counterpart across various maximum sequence lengths on the BABEL-Grounding dataset. Specifically, the Transformer baseline consists of a transformer encoder for encoding the motion sequence and a cross-attention module for injecting textual conditions. In our experiments, we configure the model with the following parameters: $d_{model}=256$, $d_{feedforward}=2048$, and $nhead=8$. The motion encoder contains 6 layers, while the cross-attention module contains 3 layers. Here, the feature dimension $d_{model}=256$ is the same as in TM-Mamba. As shown, TM-Mamba exhibits superior performance to temporal Transformer on shorter motion sequence. Meanwhile, the Transformer model runs out of GPU memory at sequence lengths exceeding 1000 due to its quadratic memory consumption. In contrast, TM-Mamba, benefiting from its linear memory cost, manages longer sequences effectively. It can be observed that increasing sequence length complicates the extraction of temporal global context, leading to a decline in motion grounding performance.
To ensure the performance gains are not due to more parameters, we compared the models’ parameter counts. The number of parameters for each model is as follows (the CLIP text encoder \textit{clip-vit-base-patch32} is excluded from the parameter count):
\begin{itemize}
    \item Transformer: 13.72M parameters
    \item TM-Mamba: 6.25M parameters
\end{itemize}
Therefore, the performance improvement cannot be attributed to a larger parameter size. 

\begin{table}[t]
\caption{Impact of the number of TM-Mamba blocks on the grounding performance. GPU memory overflow occurred with 4 blocks.
} 
	{\large
		\centering
			\resizebox{0.44\textwidth}{!}{
				\begin{tabular}{c|cccccccc}
					\toprule
					\multirow{2}{*}{\textbf{Blocks}} & \multicolumn{8}{c}{\textbf{mAP@IoU (\%)}} \\
					& 0.1   & 0.2 & 0.3 & 0.4 & 0.5 & 0.6 & 0.7 & Avg\\ \midrule \midrule
     
1  & 47.8 & 44.1 & 40.3 & 36.4 & 32.9 & 28.2 & 22.8 & 36.1 \\
2  & 53.6 & 50.0 & 46.0  & 42.0 & 37.5 & 32.0 & 25.5 & 40.9\\
3  & 53.9  & 50.5 & 46.7 & 42.8 & 38.4 & 32.6 & 26.0 & 41.6 \\
  \toprule
				\end{tabular}
			}

		\label{tab:block}
	}

\end{table}

\begin{figure*}[t]
\begin{center}
\includegraphics[width=0.8\linewidth]{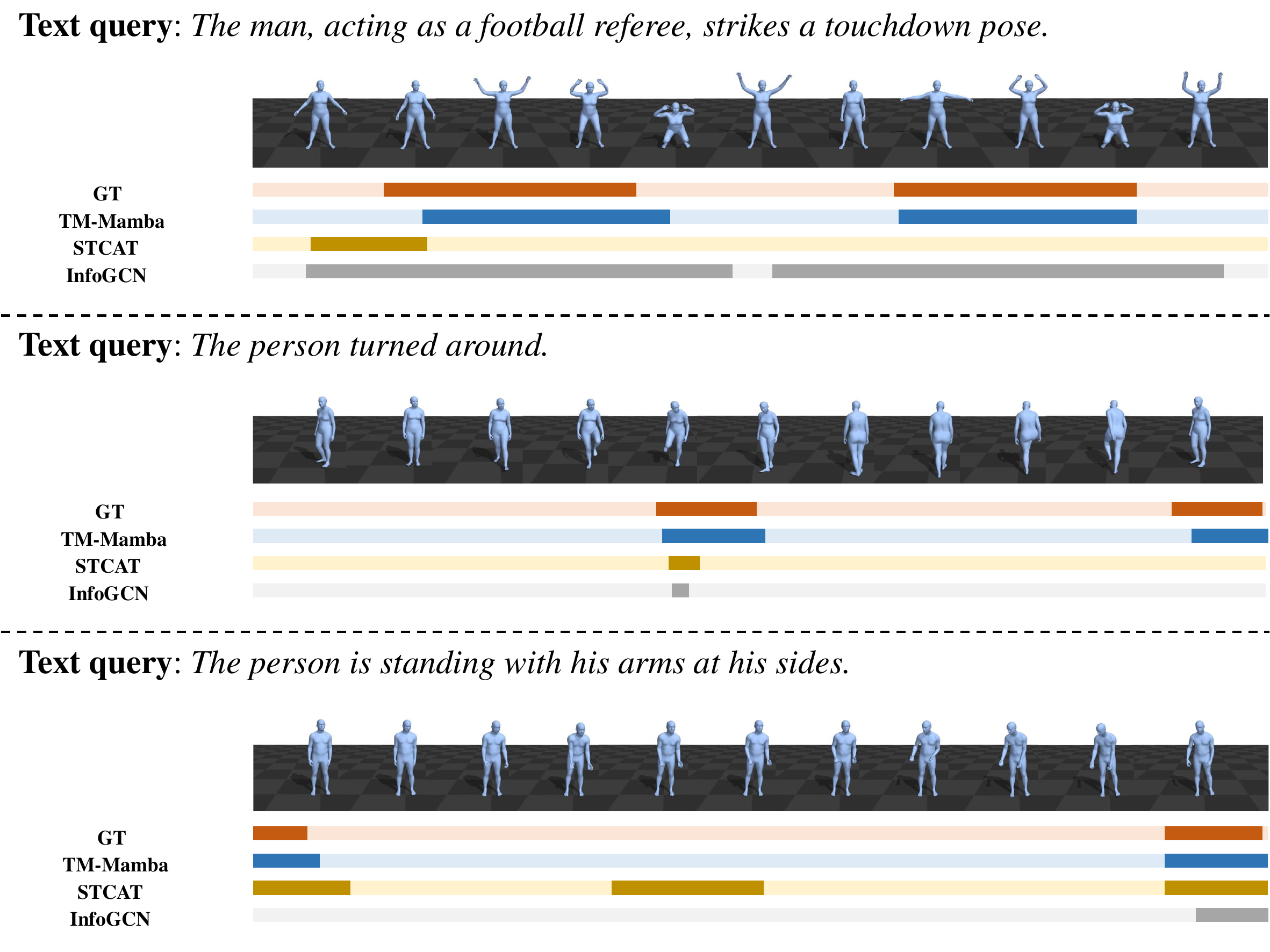}
\end{center}
   \caption{\small Visualizations of the grounding results of TM-Mamba compared to baseline models. `GT' denotes ground-truth temporal segments corresponding to the text query. Best viewed in color.}
\label{fig:supp_visualization}
\end{figure*}

\paragraph{Visualizations}
In order to demonstrate the effectiveness of our proposed model, we provide a series of visualizations on the BABEL-Grounding dataset. Figure~\ref{fig:frame} presents the frame activation scores and the ground-truth temporal segments corresponding to the text query to showcase the comparisons between the TM-Mamba and two ablative models (w/o text control and w/o relational embedding). As illustrated, our full model manifests higher activation scores within the ground-truth temporal segments in contrast to other models, thereby resulting in enhanced accuracy in the grounding performance. Figure~\ref{fig:visualization} and Figure~\ref{fig:supp_visualization} further visualizes the grounding results predicted by TM-Mamba in comparison with ablative and baseline models respectively.
As depicted, the comparative models struggle to accurately predict the time spans corresponding to the text query, while TM-Mamba achieves significantly improved grounding precision.

\paragraph{Number of Blocks}
Table~\ref{tab:block} illustrates the influence of the number of TM-Mamba blocks on grounding performance. Notably, performance begins to saturate when the number reaches 3 blocks, and encounter a GPU memory overflow at 4 blocks. 
We set the number of blocks as 3 to strike a balance between performance and computational efficiency.

\section{Conclusion}
This work introduces a novel task called text-based human motion grounding (THMG), which aims to determine the start and end timestamps of all segments from an untrimmed motion sequence given a textual description. The key challenge lies in extracting global temporal information from lengthy untrimmed sequences based on text query, while Transformer-based methods suffer from quadratic memory cost. To this end, we draw inspiration from recent advances in state space models, and propose a unified framework called TM-Mamba with linear memory cost. TM-Mamba incorporates a novel text-controlled selection mechanism into the Mamba algorithm, enabling the model to dynamically propagate input information based on text queries and extract relevant global context. A relational embedding is incorporated to model the graph topology of the human skeleton. For evaluation, a text-motion dataset called BABEL-Grounding is constructed, which is the first one that provides detailed textual descriptions with their corresponding temporal segments annotation. Rigorous experiments validate the effectiveness of the proposed model.


%

\bibliographystyle{IEEEtran}
\bibliography{ieeetip}

\begin{thebibliography}{100}
\providecommand{\url}[1]{#1}
\csname url@samestyle\endcsname
\providecommand{\newblock}{\relax}
\providecommand{\bibinfo}[2]{#2}
\providecommand{\BIBentrySTDinterwordspacing}{\spaceskip=0pt\relax}
\providecommand{\BIBentryALTinterwordstretchfactor}{4}
\providecommand{\BIBentryALTinterwordspacing}{\spaceskip=\fontdimen2\font plus
\BIBentryALTinterwordstretchfactor\fontdimen3\font minus \fontdimen4\font\relax}
\providecommand{\BIBforeignlanguage}[2]{{%
\expandafter\ifx\csname l@#1\endcsname\relax
\typeout{** WARNING: IEEEtran.bst: No hyphenation pattern has been}%
\typeout{** loaded for the language `#1'. Using the pattern for}%
\typeout{** the default language instead.}%
\else
\language=\csname l@#1\endcsname
\fi
#2}}
\providecommand{\BIBdecl}{\relax}
\BIBdecl

\bibitem{mahmood2019amass}
N.~Mahmood, N.~Ghorbani, N.~F. Troje, G.~Pons-Moll, and M.~J. Black, ``Amass: Archive of motion capture as surface shapes,'' in \emph{Proceedings of the IEEE/CVF international conference on computer vision}, 2019, pp. 5442--5451.

\bibitem{shahroudy2016ntu}
A.~Shahroudy, J.~Liu, T.-T. Ng, and G.~Wang, ``Ntu rgb+ d: A large scale dataset for 3d human activity analysis,'' in \emph{Proceedings of the IEEE conference on computer vision and pattern recognition}, 2016, pp. 1010--1019.

\bibitem{mandery2015kit}
C.~Mandery, {\"O}.~Terlemez, M.~Do, N.~Vahrenkamp, and T.~Asfour, ``The kit whole-body human motion database,'' in \emph{2015 International Conference on Advanced Robotics (ICAR)}.\hskip 1em plus 0.5em minus 0.4em\relax IEEE, 2015, pp. 329--336.

\bibitem{ionescu2013human3}
C.~Ionescu, D.~Papava, V.~Olaru, and C.~Sminchisescu, ``Human3. 6m: Large scale datasets and predictive methods for 3d human sensing in natural environments,'' \emph{IEEE transactions on pattern analysis and machine intelligence}, vol.~36, no.~7, pp. 1325--1339, 2013.

\bibitem{li2021multiscale}
M.~Li, S.~Chen, Y.~Zhao, Y.~Zhang, Y.~Wang, and Q.~Tian, ``Multiscale spatio-temporal graph neural networks for 3d skeleton-based motion prediction,'' \emph{IEEE Transactions on Image Processing}, vol.~30, pp. 7760--7775, 2021.

\bibitem{li2020multitask}
B.~Li, J.~Tian, Z.~Zhang, H.~Feng, and X.~Li, ``Multitask non-autoregressive model for human motion prediction,'' \emph{IEEE Transactions on Image Processing}, vol.~30, pp. 2562--2574, 2020.

\bibitem{wang2021pvred}
H.~Wang, J.~Dong, B.~Cheng, and J.~Feng, ``Pvred: A position-velocity recurrent encoder-decoder for human motion prediction,'' \emph{IEEE Transactions on Image Processing}, vol.~30, pp. 6096--6106, 2021.

\bibitem{wang2023dynamic}
X.~Wang, W.~Zhang, C.~Wang, Y.~Gao, and M.~Liu, ``Dynamic dense graph convolutional network for skeleton-based human motion prediction,'' \emph{IEEE Transactions on Image Processing}, vol.~33, pp. 1--15, 2023.

\bibitem{guo2023b2c}
F.~Guo, T.~Jin, S.~Zhu, X.~Xi, W.~Wang, Q.~Meng, W.~Song, and J.~Zhu, ``B2c-afm: Bi-directional co-temporal and cross-spatial attention fusion model for human action recognition,'' \emph{IEEE Transactions on Image Processing}, 2023.

\bibitem{wang2023neural}
X.~Wang, X.~Xu, and Y.~Mu, ``Neural koopman pooling: Control-inspired temporal dynamics encoding for skeleton-based action recognition,'' in \emph{Proceedings of the IEEE/CVF Conference on Computer Vision and Pattern Recognition}, 2023, pp. 10\,597--10\,607.

\bibitem{wang2024localized}
X.~Wang and Y.~Mu, ``Localized linear temporal dynamics for self-supervised skeleton action recognition,'' \emph{IEEE Transactions on Multimedia}, 2024.

\bibitem{myung2024degcn}
W.~Myung, N.~Su, J.-H. Xue, and G.~Wang, ``Degcn: Deformable graph convolutional networks for skeleton-based action recognition,'' \emph{IEEE Transactions on Image Processing}, vol.~33, pp. 2477--2490, 2024.

\bibitem{li2022smam}
Z.~Li, X.~Gong, R.~Song, P.~Duan, J.~Liu, and W.~Zhang, ``Smam: Self and mutual adaptive matching for skeleton-based few-shot action recognition,'' \emph{IEEE Transactions on Image Processing}, vol.~32, pp. 392--402, 2022.

\bibitem{zhu2022multilevel}
Y.~Zhu, H.~Shuai, G.~Liu, and Q.~Liu, ``Multilevel spatial--temporal excited graph network for skeleton-based action recognition,'' \emph{IEEE Transactions on Image Processing}, vol.~32, pp. 496--508, 2022.

\bibitem{shi2020skeleton}
L.~Shi, Y.~Zhang, J.~Cheng, and H.~Lu, ``Skeleton-based action recognition with multi-stream adaptive graph convolutional networks,'' \emph{IEEE Transactions on Image Processing}, vol.~29, pp. 9532--9545, 2020.

\bibitem{yang2021feedback}
H.~Yang, D.~Yan, L.~Zhang, Y.~Sun, D.~Li, and S.~J. Maybank, ``Feedback graph convolutional network for skeleton-based action recognition,'' \emph{IEEE Transactions on Image Processing}, vol.~31, pp. 164--175, 2021.

\bibitem{hao2021hypergraph}
X.~Hao, J.~Li, Y.~Guo, T.~Jiang, and M.~Yu, ``Hypergraph neural network for skeleton-based action recognition,'' \emph{IEEE Transactions on image processing}, vol.~30, pp. 2263--2275, 2021.

\bibitem{bian2021structural}
C.~Bian, W.~Feng, L.~Wan, and S.~Wang, ``Structural knowledge distillation for efficient skeleton-based action recognition,'' \emph{IEEE Transactions on Image Processing}, vol.~30, pp. 2963--2976, 2021.

\bibitem{cheng2021extremely}
K.~Cheng, Y.~Zhang, X.~He, J.~Cheng, and H.~Lu, ``Extremely lightweight skeleton-based action recognition with shiftgcn++,'' \emph{IEEE Transactions on Image Processing}, vol.~30, pp. 7333--7348, 2021.

\bibitem{wang2022contrast}
P.~Wang, J.~Wen, C.~Si, Y.~Qian, and L.~Wang, ``Contrast-reconstruction representation learning for self-supervised skeleton-based action recognition,'' \emph{IEEE Transactions on Image Processing}, vol.~31, pp. 6224--6238, 2022.

\bibitem{xia2017human}
G.~Xia, H.~Sun, L.~Feng, G.~Zhang, and Y.~Liu, ``Human motion segmentation via robust kernel sparse subspace clustering,'' \emph{IEEE Transactions on Image Processing}, vol.~27, no.~1, pp. 135--150, 2017.

\bibitem{yang2023lac}
D.~Yang, Y.~Wang, A.~Dantcheva, Q.~Kong, L.~Garattoni, G.~Francesca, and F.~Bremond, ``Lac-latent action composition for skeleton-based action segmentation,'' in \emph{Proceedings of the IEEE/CVF International Conference on Computer Vision}, 2023, pp. 13\,679--13\,690.

\bibitem{plappert2016kit}
M.~Plappert, C.~Mandery, and T.~Asfour, ``The kit motion-language dataset,'' \emph{Big data}, vol.~4, no.~4, pp. 236--252, 2016.

\bibitem{guo2022generating}
C.~Guo, S.~Zou, X.~Zuo, S.~Wang, W.~Ji, X.~Li, and L.~Cheng, ``Generating diverse and natural 3d human motions from text,'' in \emph{Proceedings of the IEEE/CVF Conference on Computer Vision and Pattern Recognition}, 2022, pp. 5152--5161.

\bibitem{liang2023intergen}
H.~Liang, W.~Zhang, W.~Li, J.~Yu, and L.~Xu, ``Intergen: Diffusion-based multi-human motion generation under complex interactions,'' \emph{International Journal of Computer Vision}, pp. 1--21, 2024.

\bibitem{han2023amd}
B.~Han, H.~Peng, M.~Dong, Y.~Ren, Y.~Shen, and C.~Xu, ``Amd: Autoregressive motion diffusion,'' in \emph{Proceedings of the AAAI Conference on Artificial Intelligence}, vol.~38, no.~3, 2024, pp. 2022--2030.

\bibitem{tang2023flag3d}
Y.~Tang, J.~Liu, A.~Liu, B.~Yang, W.~Dai, Y.~Rao, J.~Lu, J.~Zhou, and X.~Li, ``Flag3d: A 3d fitness activity dataset with language instruction,'' in \emph{Proceedings of the IEEE/CVF Conference on Computer Vision and Pattern Recognition}, 2023, pp. 22\,106--22\,117.

\bibitem{li2023sequential}
S.~Li, S.~Zhuang, W.~Song, X.~Zhang, H.~Chen, and A.~Hao, ``Sequential texts driven cohesive motions synthesis with natural transitions,'' in \emph{Proceedings of the IEEE/CVF International Conference on Computer Vision}, 2023, pp. 9498--9508.

\bibitem{tevet2022motionclip}
G.~Tevet, B.~Gordon, A.~Hertz, A.~H. Bermano, and D.~Cohen-Or, ``Motionclip: Exposing human motion generation to clip space,'' in \emph{European Conference on Computer Vision}.\hskip 1em plus 0.5em minus 0.4em\relax Springer, 2022, pp. 358--374.

\bibitem{petrovich2022temos}
M.~Petrovich, M.~J. Black, and G.~Varol, ``Temos: Generating diverse human motions from textual descriptions,'' in \emph{European Conference on Computer Vision}.\hskip 1em plus 0.5em minus 0.4em\relax Springer, 2022, pp. 480--497.

\bibitem{zhang2023t2m}
J.~Zhang, Y.~Zhang, X.~Cun, Y.~Zhang, H.~Zhao, H.~Lu, X.~Shen, and Y.~Shan, ``Generating human motion from textual descriptions with discrete representations,'' in \emph{Proceedings of the IEEE/CVF Conference on Computer Vision and Pattern Recognition}, 2023, pp. 14\,730--14\,740.

\bibitem{zhang2024motiondiffuse}
M.~Zhang, Z.~Cai, L.~Pan, F.~Hong, X.~Guo, L.~Yang, and Z.~Liu, ``Motiondiffuse: Text-driven human motion generation with diffusion model,'' \emph{IEEE Transactions on Pattern Analysis and Machine Intelligence}, 2024.

\bibitem{guo2022tm2t}
C.~Guo, X.~Zuo, S.~Wang, and L.~Cheng, ``Tm2t: Stochastic and tokenized modeling for the reciprocal generation of 3d human motions and texts,'' in \emph{European Conference on Computer Vision}.\hskip 1em plus 0.5em minus 0.4em\relax Springer, 2022, pp. 580--597.

\bibitem{jiang2023motiongpt}
B.~Jiang, X.~Chen, W.~Liu, J.~Yu, G.~Yu, and T.~Chen, ``Motiongpt: Human motion as a foreign language,'' \emph{Advances in Neural Information Processing Systems}, vol.~36, 2024.

\bibitem{goel2023iterative}
P.~Goel, K.-C. Wang, C.~K. Liu, and K.~Fatahalian, ``Iterative motion editing with natural language,'' in \emph{ACM SIGGRAPH 2024 Conference Papers}, 2024, pp. 1--9.

\bibitem{kim2023flame}
J.~Kim, J.~Kim, and S.~Choi, ``Flame: Free-form language-based motion synthesis \& editing,'' in \emph{Proceedings of the AAAI Conference on Artificial Intelligence}, vol.~37, no.~7, 2023, pp. 8255--8263.

\bibitem{zhang2023finemogen}
M.~Zhang, H.~Li, Z.~Cai, J.~Ren, L.~Yang, and Z.~Liu, ``Finemogen: Fine-grained spatio-temporal motion generation and editing,'' \emph{Advances in Neural Information Processing Systems}, vol.~36, 2024.

\bibitem{endo2023motion}
M.~Endo, J.~Hsu, J.~Li, and J.~Wu, ``Motion question answering via modular motion programs,'' in \emph{International Conference on Machine Learning}.\hskip 1em plus 0.5em minus 0.4em\relax PMLR, 2023, pp. 9312--9328.

\bibitem{lei2021detecting}
J.~Lei, T.~L. Berg, and M.~Bansal, ``Detecting moments and highlights in videos via natural language queries,'' \emph{Advances in Neural Information Processing Systems}, vol.~34, pp. 11\,846--11\,858, 2021.

\bibitem{yang2022tubedetr}
A.~Yang, A.~Miech, J.~Sivic, I.~Laptev, and C.~Schmid, ``Tubedetr: Spatio-temporal video grounding with transformers,'' in \emph{Proceedings of the IEEE/CVF Conference on Computer Vision and Pattern Recognition}, 2022, pp. 16\,442--16\,453.

\bibitem{lin2023univtg}
K.~Q. Lin, P.~Zhang, J.~Chen, S.~Pramanick, D.~Gao, A.~J. Wang, R.~Yan, and M.~Z. Shou, ``Univtg: Towards unified video-language temporal grounding,'' in \emph{Proceedings of the IEEE/CVF International Conference on Computer Vision}, 2023, pp. 2794--2804.

\bibitem{mun2020local}
J.~Mun, M.~Cho, and B.~Han, ``Local-global video-text interactions for temporal grounding,'' in \emph{Proceedings of the IEEE/CVF Conference on Computer Vision and Pattern Recognition}, 2020, pp. 10\,810--10\,819.

\bibitem{zhang2020does}
Z.~Zhang, Z.~Zhao, Y.~Zhao, Q.~Wang, H.~Liu, and L.~Gao, ``Where does it exist: Spatio-temporal video grounding for multi-form sentences,'' in \emph{Proceedings of the IEEE/CVF Conference on Computer Vision and Pattern Recognition}, 2020, pp. 10\,668--10\,677.

\bibitem{paul2018w}
S.~Paul, S.~Roy, and A.~K. Roy-Chowdhury, ``W-talc: Weakly-supervised temporal activity localization and classification,'' in \emph{Proceedings of the European conference on computer vision (ECCV)}, 2018, pp. 563--579.

\bibitem{liu2019completeness}
D.~Liu, T.~Jiang, and Y.~Wang, ``Completeness modeling and context separation for weakly supervised temporal action localization,'' in \emph{Proceedings of the IEEE/CVF conference on computer vision and pattern recognition}, 2019, pp. 1298--1307.

\bibitem{shou2017cdc}
Z.~Shou, J.~Chan, A.~Zareian, K.~Miyazawa, and S.-F. Chang, ``Cdc: Convolutional-de-convolutional networks for precise temporal action localization in untrimmed videos,'' in \emph{Proceedings of the IEEE conference on computer vision and pattern recognition}, 2017, pp. 5734--5743.

\bibitem{liu2024stepwise}
M.~Liu, L.~Wang, S.~Zhou, K.~Xia, Q.~Wu, Q.~Zhang, and G.~Hua, ``Stepwise multi-grained boundary detector for point-supervised temporal action localization,'' in \emph{European Conference on Computer Vision}.\hskip 1em plus 0.5em minus 0.4em\relax Springer, 2024, pp. 333--349.

\bibitem{zhang2022temporal}
Z.~Zhang and J.~Yang, ``Temporal sentiment localization: Listen and look in untrimmed videos,'' in \emph{Proceedings of the 30th ACM International Conference on Multimedia}, 2022, pp. 199--208.

\bibitem{zhang2022actionformer}
C.-L. Zhang, J.~Wu, and Y.~Li, ``Actionformer: Localizing moments of actions with transformers,'' in \emph{European Conference on Computer Vision}.\hskip 1em plus 0.5em minus 0.4em\relax Springer, 2022, pp. 492--510.

\bibitem{xu2021videoclip}
H.~Xu, G.~Ghosh, P.-Y. Huang, D.~Okhonko, A.~Aghajanyan, F.~Metze, L.~Zettlemoyer, and C.~Feichtenhofer, ``Videoclip: Contrastive pre-training for zero-shot video-text understanding,'' in \emph{Proceedings of the 2021 Conference on Empirical Methods in Natural Language Processing}, 2021, pp. 6787--6800.

\bibitem{yu2023frame}
Q.~Yu and K.~Fujiwara, ``Frame-level label refinement for skeleton-based weakly-supervised action recognition,'' in \emph{Proceedings of the AAAI Conference on Artificial Intelligence}, vol.~37, no.~3, 2023, pp. 3322--3330.

\bibitem{guo2020action2motion}
C.~Guo, X.~Zuo, S.~Wang, S.~Zou, Q.~Sun, A.~Deng, M.~Gong, and L.~Cheng, ``Action2motion: Conditioned generation of 3d human motions,'' in \emph{Proceedings of the 28th ACM International Conference on Multimedia}, 2020, pp. 2021--2029.

\bibitem{wang2022humanise}
Z.~Wang, Y.~Chen, T.~Liu, Y.~Zhu, W.~Liang, and S.~Huang, ``Humanise: Language-conditioned human motion generation in 3d scenes,'' \emph{Advances in Neural Information Processing Systems}, vol.~35, pp. 14\,959--14\,971, 2022.

\bibitem{zhou2023avatargpt}
Z.~Zhou, Y.~Wan, and B.~Wang, ``Avatargpt: All-in-one framework for motion understanding planning generation and beyond,'' in \emph{Proceedings of the IEEE/CVF Conference on Computer Vision and Pattern Recognition}, 2024, pp. 1357--1366.

\bibitem{zhong2023attt2m}
C.~Zhong, L.~Hu, Z.~Zhang, and S.~Xia, ``Attt2m: Text-driven human motion generation with multi-perspective attention mechanism,'' in \emph{Proceedings of the IEEE/CVF International Conference on Computer Vision}, 2023, pp. 509--519.

\bibitem{zhai2023language}
Y.~Zhai, M.~Huang, T.~Luan, L.~Dong, I.~Nwogu, S.~Lyu, D.~Doermann, and J.~Yuan, ``Language-guided human motion synthesis with atomic actions,'' in \emph{Proceedings of the 31st ACM International Conference on Multimedia}, 2023, pp. 5262--5271.

\bibitem{gu2023mamba}
A.~Gu and T.~Dao, ``Mamba: Linear-time sequence modeling with selective state spaces,'' in \emph{First Conference on Language Modeling}, 2024.

\bibitem{qiao2024vl}
Y.~Qiao, Z.~Yu, L.~Guo, S.~Chen, Z.~Zhao, M.~Sun, Q.~Wu, and J.~Liu, ``Vl-mamba: Exploring state space models for multimodal learning,'' \emph{arXiv preprint arXiv:2403.13600}, 2024.

\bibitem{zhao2024cobra}
H.~Zhao, M.~Zhang, W.~Zhao, P.~Ding, S.~Huang, and D.~Wang, ``Cobra: Extending mamba to multi-modal large language model for efficient inference,'' in \emph{Proceedings of the AAAI Conference on Artificial Intelligence}, vol.~39, no.~10, 2025, pp. 10\,421--10\,429.

\bibitem{punnakkal2021babel}
A.~R. Punnakkal, A.~Chandrasekaran, N.~Athanasiou, A.~Quiros-Ramirez, and M.~J. Black, ``Babel: Bodies, action and behavior with english labels,'' in \emph{Proceedings of the IEEE/CVF Conference on Computer Vision and Pattern Recognition}, 2021, pp. 722--731.

\bibitem{lin2023motion}
J.~Lin, A.~Zeng, S.~Lu, Y.~Cai, R.~Zhang, H.~Wang, and L.~Zhang, ``Motion-x: A large-scale 3d expressive whole-body human motion dataset,'' \emph{Advances in Neural Information Processing Systems}, vol.~36, 2024.

\bibitem{peng2023hoi}
X.~Peng, Y.~Xie, Z.~Wu, V.~Jampani, D.~Sun, and H.~Jiang, ``Hoi-diff: Text-driven synthesis of 3d human-object interactions using diffusion models,'' in \emph{Proceedings of the Computer Vision and Pattern Recognition Conference}, 2025, pp. 2878--2888.

\bibitem{delmas2022posescript}
G.~Delmas, P.~Weinzaepfel, T.~Lucas, F.~Moreno-Noguer, and G.~Rogez, ``Posescript: 3d human poses from natural language,'' in \emph{European Conference on Computer Vision}.\hskip 1em plus 0.5em minus 0.4em\relax Springer, 2022, pp. 346--362.

\bibitem{delmas2023posefix}
G.~Delmas, P.~Weinzaepfel, F.~Moreno-Noguer, and G.~Rogez, ``Posefix: Correcting 3d human poses with natural language,'' in \emph{Proceedings of the IEEE/CVF International Conference on Computer Vision}, 2023, pp. 15\,018--15\,028.

\bibitem{cai2022humman}
Z.~Cai, D.~Ren, A.~Zeng, Z.~Lin, T.~Yu, W.~Wang, X.~Fan, Y.~Gao, Y.~Yu, L.~Pan \emph{et~al.}, ``Humman: Multi-modal 4d human dataset for versatile sensing and modeling,'' in \emph{European Conference on Computer Vision}.\hskip 1em plus 0.5em minus 0.4em\relax Springer, 2022, pp. 557--577.

\bibitem{ahuja2019language2pose}
C.~Ahuja and L.-P. Morency, ``Language2pose: Natural language grounded pose forecasting,'' in \emph{2019 International Conference on 3D Vision (3DV)}.\hskip 1em plus 0.5em minus 0.4em\relax IEEE, 2019, pp. 719--728.

\bibitem{ghosh2021synthesis}
A.~Ghosh, N.~Cheema, C.~Oguz, C.~Theobalt, and P.~Slusallek, ``Synthesis of compositional animations from textual descriptions,'' in \emph{Proceedings of the IEEE/CVF international conference on computer vision}, 2021, pp. 1396--1406.

\bibitem{yuan2023physdiff}
Y.~Yuan, J.~Song, U.~Iqbal, A.~Vahdat, and J.~Kautz, ``Physdiff: Physics-guided human motion diffusion model,'' in \emph{Proceedings of the IEEE/CVF International Conference on Computer Vision}, 2023, pp. 16\,010--16\,021.

\bibitem{ma2022pretrained}
J.~Ma, S.~Bai, and C.~Zhou, ``Pretrained diffusion models for unified human motion synthesis,'' \emph{arXiv preprint arXiv:2212.02837}, 2022.

\bibitem{jin2023act}
P.~Jin, Y.~Wu, Y.~Fan, Z.~Sun, W.~Yang, and L.~Yuan, ``Act as you wish: Fine-grained control of motion diffusion model with hierarchical semantic graphs,'' \emph{Advances in Neural Information Processing Systems}, vol.~36, 2024.

\bibitem{azadi2023make}
S.~Azadi, A.~Shah, T.~Hayes, D.~Parikh, and S.~Gupta, ``Make-an-animation: Large-scale text-conditional 3d human motion generation,'' in \emph{Proceedings of the IEEE/CVF International Conference on Computer Vision}, 2023, pp. 15\,039--15\,048.

\bibitem{wang2023fg}
Y.~Wang, Z.~Leng, F.~W. Li, S.-C. Wu, and X.~Liang, ``Fg-t2m: Fine-grained text-driven human motion generation via diffusion model,'' in \emph{Proceedings of the IEEE/CVF International Conference on Computer Vision}, 2023, pp. 22\,035--22\,044.

\bibitem{kong2023priority}
H.~Kong, K.~Gong, D.~Lian, M.~B. Mi, and X.~Wang, ``Priority-centric human motion generation in discrete latent space,'' in \emph{Proceedings of the IEEE/CVF International Conference on Computer Vision}, 2023, pp. 14\,806--14\,816.

\bibitem{karunratanakul2023guided}
K.~Karunratanakul, K.~Preechakul, S.~Suwajanakorn, and S.~Tang, ``Guided motion diffusion for controllable human motion synthesis,'' in \emph{Proceedings of the IEEE/CVF International Conference on Computer Vision}, 2023, pp. 2151--2162.

\bibitem{qian2023breaking}
Y.~Qian, J.~Urbanek, A.~G. Hauptmann, and J.~Won, ``Breaking the limits of text-conditioned 3d motion synthesis with elaborative descriptions,'' in \emph{Proceedings of the IEEE/CVF International Conference on Computer Vision}, 2023, pp. 2306--2316.

\bibitem{chen2023executing}
X.~Chen, B.~Jiang, W.~Liu, Z.~Huang, B.~Fu, T.~Chen, and G.~Yu, ``Executing your commands via motion diffusion in latent space,'' in \emph{Proceedings of the IEEE/CVF Conference on Computer Vision and Pattern Recognition}, 2023, pp. 18\,000--18\,010.

\bibitem{zhou2023ude}
Z.~Zhou and B.~Wang, ``Ude: A unified driving engine for human motion generation,'' in \emph{Proceedings of the IEEE/CVF Conference on Computer Vision and Pattern Recognition}, 2023, pp. 5632--5641.

\bibitem{lin2023being}
J.~Lin, J.~Chang, L.~Liu, G.~Li, L.~Lin, Q.~Tian, and C.-w. Chen, ``Being comes from not-being: Open-vocabulary text-to-motion generation with wordless training,'' in \emph{Proceedings of the IEEE/CVF conference on computer vision and pattern recognition}, 2023, pp. 23\,222--23\,231.

\bibitem{yang2023synthesizing}
Z.~Yang, B.~Su, and J.-R. Wen, ``Synthesizing long-term human motions with diffusion models via coherent sampling,'' in \emph{Proceedings of the 31st ACM International Conference on Multimedia}, 2023, pp. 3954--3964.

\bibitem{han2023hutumotion}
G.~Han, S.~Huang, M.~Gong, and J.~Tang, ``Hutumotion: Human-tuned navigation of latent motion diffusion models with minimal feedback,'' in \emph{Proceedings of the AAAI Conference on Artificial Intelligence}, vol.~38, no.~3, 2024, pp. 2031--2039.

\bibitem{hoang2024motionmix}
N.~Hoang, K.~Gong, C.~Guo, and M.~Mi, ``Motionmix: Weakly-supervised diffusion for controllable motion generation,'' \emph{Proceedings of the AAAI Conference on Artificial Intelligence}, vol.~38, pp. 2157--2165, 03 2024.

\bibitem{xie2023towards}
Z.~Xie, Y.~Wu, X.~Gao, Z.~Sun, W.~Yang, and X.~Liang, ``Towards detailed text-to-motion synthesis via basic-to-advanced hierarchical diffusion model,'' in \emph{Proceedings of the AAAI Conference on Artificial Intelligence}, vol.~38, no.~6, 2024, pp. 6252--6260.

\bibitem{zhang2023motiongpt}
Y.~Zhang, D.~Huang, B.~Liu, S.~Tang, Y.~Lu, L.~Chen, L.~Bai, Q.~Chu, N.~Yu, and W.~Ouyang, ``Motiongpt: Finetuned llms are general-purpose motion generators,'' in \emph{Proceedings of the AAAI Conference on Artificial Intelligence}, vol.~38, no.~7, 2024, pp. 7368--7376.

\bibitem{shafir2023human}
Y.~Shafir, G.~Tevet, R.~Kapon, and A.~H. Bermano, ``Human motion diffusion as a generative prior,'' \emph{arXiv preprint arXiv:2303.01418}, 2023.

\bibitem{wei2023nerm}
D.~Wei, H.~Sun, B.~Li, X.~Sun, S.~Hu, W.~Li, and J.~Lu, ``Nerm: Learning neural representations for high-framerate human motion synthesis,'' in \emph{The Twelfth International Conference on Learning Representations}, 2023.

\bibitem{xie2023omnicontrol}
Y.~Xie, V.~Jampani, L.~Zhong, D.~Sun, and H.~Jiang, ``Omnicontrol: Control any joint at any time for human motion generation,'' in \emph{The Twelfth International Conference on Learning Representations}, 2024.

\bibitem{athanasiou2022teach}
N.~Athanasiou, M.~Petrovich, M.~J. Black, and G.~Varol, ``Teach: Temporal action composition for 3d humans,'' in \emph{2022 International Conference on 3D Vision (3DV)}.\hskip 1em plus 0.5em minus 0.4em\relax IEEE, 2022, pp. 414--423.

\bibitem{dabral2023mofusion}
R.~Dabral, M.~H. Mughal, V.~Golyanik, and C.~Theobalt, ``Mofusion: A framework for denoising-diffusion-based motion synthesis,'' in \emph{Proceedings of the IEEE/CVF Conference on Computer Vision and Pattern Recognition}, 2023, pp. 9760--9770.

\bibitem{petrovich2023tmr}
M.~Petrovich, M.~J. Black, and G.~Varol, ``Tmr: Text-to-motion retrieval using contrastive 3d human motion synthesis,'' in \emph{Proceedings of the IEEE/CVF International Conference on Computer Vision}, 2023, pp. 9488--9497.

\bibitem{messina2023text}
N.~Messina, J.~Sedmidubsky, F.~Falchi, and T.~Rebok, ``Text-to-motion retrieval: Towards joint understanding of human motion data and natural language,'' in \emph{Proceedings of the 46th International ACM SIGIR Conference on Research and Development in Information Retrieval}, 2023, pp. 2420--2425.

\bibitem{gu2021efficiently}
A.~Gu, K.~Goel, and C.~R{\'e}, ``Efficiently modeling long sequences with structured state spaces,'' \emph{arXiv preprint arXiv:2111.00396}, 2021.

\bibitem{gu2020hippo}
A.~Gu, T.~Dao, S.~Ermon, A.~Rudra, and C.~R{\'e}, ``Hippo: Recurrent memory with optimal polynomial projections,'' \emph{Advances in neural information processing systems}, vol.~33, pp. 1474--1487, 2020.

\bibitem{smith2022simplified}
J.~T. Smith, A.~Warrington, and S.~Linderman, ``Simplified state space layers for sequence modeling,'' in \emph{The Eleventh International Conference on Learning Representations}, 2023.

\bibitem{fu2022hungry}
T.~Dao, D.~Y. Fu, K.~K. Saab, A.~W. Thomas, A.~Rudra, and C.~R{\'e}, ``Hungry hungry hippos: Towards language modeling with state space models,'' in \emph{Proceedings of the 11th International Conference on Learning Representations (ICLR)}, 2023.

\bibitem{mehta2022long}
H.~Mehta, A.~Gupta, A.~Cutkosky, and B.~Neyshabur, ``Long range language modeling via gated state spaces,'' \emph{arXiv preprint arXiv:2206.13947}, 2022.

\bibitem{gu2022parameterization}
A.~Gu, K.~Goel, A.~Gupta, and C.~R{\'e}, ``On the parameterization and initialization of diagonal state space models,'' \emph{Advances in Neural Information Processing Systems}, vol.~35, pp. 35\,971--35\,983, 2022.

\bibitem{gupta2022diagonal}
A.~Gupta, A.~Gu, and J.~Berant, ``Diagonal state spaces are as effective as structured state spaces,'' \emph{Advances in Neural Information Processing Systems}, vol.~35, pp. 22\,982--22\,994, 2022.

\bibitem{hasani2022liquid}
R.~Hasani, M.~Lechner, T.-H. Wang, M.~Chahine, A.~Amini, and D.~Rus, ``Liquid structural state-space models,'' \emph{arXiv preprint arXiv:2209.12951}, 2022.

\bibitem{gu2022train}
A.~Gu, I.~Johnson, A.~Timalsina, A.~Rudra, and C.~R{\'e}, ``How to train your hippo: State space models with generalized orthogonal basis projections,'' in \emph{Proceedings of the 11th International Conference on Learning Representations (ICLR)}, 2023.

\bibitem{dao2024transformers}
T.~Dao and A.~Gu, ``Transformers are ssms: Generalized models and efficient algorithms through structured state space duality,'' \emph{Proceedings of Machine Learning Research}, vol. 235, pp. 10\,041--10\,071, 2024.

\bibitem{guo2024mambair}
H.~Guo, J.~Li, T.~Dai, Z.~Ouyang, X.~Ren, and S.-T. Xia, ``Mambair: A simple baseline for image restoration with state-space model,'' in \emph{European conference on computer vision}.\hskip 1em plus 0.5em minus 0.4em\relax Springer, 2024, pp. 222--241.

\bibitem{wang2024semi}
C.~Ma and Z.~Wang, ``Semi-mamba-unet: Pixel-level contrastive and cross-supervised visual mamba-based unet for semi-supervised medical image segmentation,'' \emph{Knowledge-Based Systems}, vol. 300, p. 112203, 2024.

\bibitem{ruan2024vm}
J.~Ruan, J.~Li, and S.~Xiang, ``Vm-unet: Vision mamba unet for medical image segmentation,'' \emph{ACM Transactions on Multimedia Computing, Communications and Applications}, 2024.

\bibitem{ma2024u}
J.~Ma, F.~Li, and B.~Wang, ``U-mamba: Enhancing long-range dependency for biomedical image segmentation,'' \emph{arXiv preprint arXiv:2401.04722}, 2024.

\bibitem{xing2024segmamba}
Z.~Xing, T.~Ye, Y.~Yang, G.~Liu, and L.~Zhu, ``Segmamba: Long-range sequential modeling mamba for 3d medical image segmentation,'' in \emph{International conference on medical image computing and computer-assisted intervention}.\hskip 1em plus 0.5em minus 0.4em\relax Springer, 2024, pp. 578--588.

\bibitem{liu2024swin}
J.~Liu, H.~Yang, H.-Y. Zhou, Y.~Xi, L.~Yu, C.~Li, Y.~Liang, G.~Shi, Y.~Yu, S.~Zhang \emph{et~al.}, ``Swin-umamba: Mamba-based unet with imagenet-based pretraining,'' in \emph{International conference on medical image computing and computer-assisted intervention}.\hskip 1em plus 0.5em minus 0.4em\relax Springer, 2024, pp. 615--625.

\bibitem{liang2024pointmamba}
D.~Liang, X.~Zhou, W.~Xu, X.~Zhu, Z.~Zou, X.~Ye, X.~Tan, and X.~Bai, ``Pointmamba: A simple state space model for point cloud analysis,'' \emph{Advances in neural information processing systems}, vol.~37, pp. 32\,653--32\,677, 2024.

\bibitem{yang2024vivim}
Y.~Yang, Z.~Xing, and L.~Zhu, ``Vivim: a video vision mamba for medical video object segmentation,'' \emph{arXiv preprint arXiv:2401.14168}, 2024.

\bibitem{li2024videomamba}
K.~Li, X.~Li, Y.~Wang, Y.~He, Y.~Wang, L.~Wang, and Y.~Qiao, ``Videomamba: State space model for efficient video understanding,'' in \emph{European conference on computer vision}.\hskip 1em plus 0.5em minus 0.4em\relax Springer, 2024, pp. 237--255.

\bibitem{chen2024video}
G.~Chen, Y.~Huang, J.~Xu, B.~Pei, Z.~Chen, Z.~Li, J.~Wang, K.~Li, T.~Lu, and L.~Wang, ``Video mamba suite: State space model as a versatile alternative for video understanding,'' \emph{arXiv preprint arXiv:2403.09626}, 2024.

\bibitem{he2024pan}
X.~He, K.~Cao, J.~Zhang, K.~Yan, Y.~Wang, R.~Li, C.~Xie, D.~Hong, and M.~Zhou, ``Pan-mamba: Effective pan-sharpening with state space model,'' \emph{Information Fusion}, vol. 115, p. 102779, 2025.

\bibitem{behrouz2024graph}
A.~Behrouz and F.~Hashemi, ``Graph mamba: Towards learning on graphs with state space models,'' \emph{arXiv preprint arXiv:2402.08678}, 2024.

\bibitem{wang2024graph}
C.~Wang, O.~Tsepa, J.~Ma, and B.~Wang, ``Graph-mamba: Towards long-range graph sequence modeling with selective state spaces,'' \emph{arXiv preprint arXiv:2402.00789}, 2024.

\bibitem{zhu2024vision}
L.~Zhu, B.~Liao, Q.~Zhang, X.~Wang, W.~Liu, and X.~Wang, ``Vision mamba: Efficient visual representation learning with bidirectional state space model,'' \emph{arXiv preprint arXiv:2401.09417}, 2024.

\bibitem{liu2024vmamba}
Y.~Liu, Y.~Tian, Y.~Zhao, H.~Yu, L.~Xie, Y.~Wang, Q.~Ye, J.~Jiao, and Y.~Liu, ``Vmamba: Visual state space model,'' \emph{Advances in neural information processing systems}, vol.~37, pp. 103\,031--103\,063, 2024.

\bibitem{zhang2024motion}
Z.~Zhang, A.~Liu, I.~Reid, R.~Hartley, B.~Zhuang, and H.~Tang, ``Motion mamba: Efficient and long sequence motion generation,'' in \emph{European Conference on Computer Vision}.\hskip 1em plus 0.5em minus 0.4em\relax Springer, 2024, pp. 265--282.

\bibitem{Qwen25VL}
S.~Bai, K.~Chen, X.~Liu, J.~Wang, W.~Ge, S.~Song, K.~Dang, P.~Wang, S.~Wang, J.~Tang, H.~Zhong, Y.~Zhu, M.~Yang, Z.~Li, J.~Wan, P.~Wang, W.~Ding, Z.~Fu, Y.~Xu, J.~Ye, X.~Zhang, T.~Xie, Z.~Cheng, H.~Zhang, Z.~Yang, H.~Xu, and J.~Lin, ``Qwen2.5-vl technical report,'' \emph{arXiv preprint arXiv:2502.13923}, 2025.

\bibitem{radford2021learning}
A.~Radford, J.~W. Kim, C.~Hallacy, A.~Ramesh, G.~Goh, S.~Agarwal, G.~Sastry, A.~Askell, P.~Mishkin, J.~Clark \emph{et~al.}, ``Learning transferable visual models from natural language supervision,'' in \emph{International conference on machine learning}.\hskip 1em plus 0.5em minus 0.4em\relax PMLR, 2021, pp. 8748--8763.

\bibitem{chen2021channel}
Y.~Chen, Z.~Zhang, C.~Yuan, B.~Li, Y.~Deng, and W.~Hu, ``Channel-wise topology refinement graph convolution for skeleton-based action recognition,'' in \emph{Proceedings of the IEEE/CVF international conference on computer vision}, 2021, pp. 13\,359--13\,368.

\bibitem{chi2022infogcn}
H.-g. Chi, M.~H. Ha, S.~Chi, S.~W. Lee, Q.~Huang, and K.~Ramani, ``Infogcn: Representation learning for human skeleton-based action recognition,'' in \emph{Proceedings of the IEEE/CVF conference on computer vision and pattern recognition}, 2022, pp. 20\,186--20\,196.

\bibitem{liu2020disentangling}
Z.~Liu, H.~Zhang, Z.~Chen, Z.~Wang, and W.~Ouyang, ``Disentangling and unifying graph convolutions for skeleton-based action recognition,'' in \emph{Proceedings of the IEEE/CVF conference on computer vision and pattern recognition}, 2020, pp. 143--152.

\bibitem{shi2019two}
L.~Shi, Y.~Zhang, J.~Cheng, and H.~Lu, ``Two-stream adaptive graph convolutional networks for skeleton-based action recognition,'' in \emph{Proceedings of the IEEE/CVF conference on computer vision and pattern recognition}, 2019, pp. 12\,026--12\,035.

\bibitem{yan2018spatial}
S.~Yan, Y.~Xiong, and D.~Lin, ``Spatial temporal graph convolutional networks for skeleton-based action recognition,'' in \emph{Proceedings of the AAAI conference on artificial intelligence}, vol.~32, no.~1, 2018.

\bibitem{loshchilov2017decoupled}
I.~Loshchilov and F.~Hutter, ``Decoupled weight decay regularization,'' \emph{arXiv preprint arXiv:1711.05101}, 2017.

\bibitem{caba2015activitynet}
F.~Caba~Heilbron, V.~Escorcia, B.~Ghanem, and J.~Carlos~Niebles, ``Activitynet: A large-scale video benchmark for human activity understanding,'' in \emph{Proceedings of the ieee conference on computer vision and pattern recognition}, 2015, pp. 961--970.

\bibitem{jang2023knowing}
J.~Jang, J.~Park, J.~Kim, H.~Kwon, and K.~Sohn, ``Knowing where to focus: Event-aware transformer for video grounding,'' in \emph{Proceedings of the IEEE/CVF International Conference on Computer Vision}, 2023, pp. 13\,846--13\,856.

\bibitem{jin2022embracing}
Y.~Jin, Z.~Yuan, Y.~Mu \emph{et~al.}, ``Embracing consistency: A one-stage approach for spatio-temporal video grounding,'' \emph{Advances in Neural Information Processing Systems}, vol.~35, pp. 29\,192--29\,204, 2022.

\bibitem{dao2022flashattention}
T.~Dao, D.~Fu, S.~Ermon, A.~Rudra, and C.~R{\'e}, ``Flashattention: Fast and memory-efficient exact attention with io-awareness,'' \emph{Advances in Neural Information Processing Systems}, vol.~35, pp. 16\,344--16\,359, 2022.

\end{thebibliography}

%

\end{document}